\documentclass[twoside,11pt]{article}
\usepackage{jair, theapa, rawfonts}
\usepackage{graphicx}
\usepackage{amsmath}
\usepackage{amssymb}
\usepackage[amsmath,amsthm,thmmarks]{ntheorem} 

\usepackage{booktabs}
\usepackage[inline]{enumitem}
\usepackage{multirow}
\usepackage{xcolor}
\usepackage{tikz}
\usetikzlibrary{shapes.geometric}
\usepackage{xspace}

\usepackage[acronym]{glossaries}

\usepackage{tabu}
\newcolumntype{M}[1]{>{\centering\arraybackslash}m{#1}}
\newcolumntype{L}[1]{>{\arraybackslash}m{#1}}

\theoremstyle{definition}
\theoremsymbol{\ensuremath{\Box}}
\newtheorem{example}{Example}
\newtheorem{definition}{Definition}
\newtheorem{namedtheorem}{Explanation}

\def\SHOIQ{\ensuremath{\mathcal{SHOIQ}}}
\def\ALC{\ensuremath{\mathcal{ALC}}}
\def\DLLite{\ensuremath{\mathrm{DL\!{-}Lite}}}
\def\coNP{\textsc{coNP}\xspace}
\def\PSPACE{\textsc{PSpace}\xspace}
\def\eg{e.g.,\xspace}
\def\ie{i.e.,\xspace}

\DeclareRobustCommand\checkmark{%
      \tikz{\fill[scale=0.4](0,.35) -- (.25,0) -- (1,.7) -- (.25,.15) -- cycle;}%
}

\newcommand{\citeauthory}[1]{\shortciteA{#1}}

\newcommand{\algo}[1]{\emph{#1}}

\newacronym{AR}{AR}{ABox Repair}
\newacronym{kb}{KB}{Knowledge Base}
\newacronym{kg}{KG}{Knowledge Graph}
\newacronym{DL}{DL}{Description Logic}
\newacronym{tgc}{TGC}{Tuple generating constraint}
\newacronym{cdd}{CDD}{}
\newacronym{gqr}{GQR}{}

\title{Dealing with Inconsistency for Reasoning over Knowledge Graphs: A Survey}
\author{%
\name Anastasios Nentidis$^1$ \email tasosnent@iit.demokritos.gr\\
\name Charilaos Akasiadis$^1$ \email cakasiadis@iit.demokritos.gr \\
\name Angelos Charalambidis$^{1,2}$ \email acharal@iit.demokritos.gr \\
\name Alexander Artikis$^{1,3}$ \email a.artikis@unipi.gr \\
\addr $^1$ National Centre of Scientific Research ``Demokritos'' \\
\addr $^2$ Harokopio University \\
\addr $^3$ University of Piraeus}

\ShortHeadings{Dealing with Inconsistency for Reasoning over KGs}
{Nentidis, Akasiadis, Charalambidis \& Artikis}
\begin{document}

\maketitle

\begin{abstract}
In Knowledge Graphs (KGs), where the schema of the data is usually defined by
particular ontologies, reasoning is a necessity to perform a range of tasks, such
as retrieval of information, question answering, and the derivation of new knowledge.    
However, information to populate KGs is often extracted (semi-) automatically
from natural language resources, or by integrating datasets that follow
different semantic schemas, resulting in KG inconsistency. This, however, hinders the
process of reasoning.  
In this survey, we focus on how to perform reasoning on inconsistent KGs, by
analyzing the state of the art towards three complementary directions: 
a) the detection of the parts of the KG that cause the inconsistency, 
b) the fixing of an inconsistent KG to render it consistent, and 
c) the inconsistency-tolerant reasoning. 
We discuss existing work from a range of relevant fields focusing on how, and
in which cases they are related to the above directions. We also highlight persisting
challenges and future directions.
\end{abstract}

\section{Introduction}
% \todo[inline]{Lead:Tasos}
% \todo[inline]{Motivation from Enexa DoA + powerful citations}
% -Intro to KGs.
\Glspl{kg} are widely adopted by world-leading organizations in a
variety of fields, including technology, banking, and media, supporting a range
of applications in the areas of knowledge acquisition, integration, and
maintenance~\cite{Abu-Salih2021}.
One of the key benefits of \glspl{kg} is the direct integration of formal reasoning.
However, in practice, this is often hindered by inconsistencies~\cite{Pensel2018}.
\glspl{kg} can be seen as special cases of \glspl{kb} where extensional knowledge, called the ABox, is organized in the form of a graph. Implicit knowledge is derived from the \gls{kg} through a set of axioms, called the TBox, typically expressed as first-order formulas.
Knowledge Graphs, a special kind of \glspl{kb}, can be considered to be augmented versions of ontologies, while in some cases the terms are used interchangeably~\shortcite{ehrlinger2016towards,mitchell2018never,heist2020knowledge}. Other works use the term ontology when referring to solely a KG's TBox~\shortcite{wang2018acekg,tran2020fast}.
\citeauthory{Hogan2022} provide a comprehensive introduction to KGs and related research directions.

% A KG $\mathcal{K} = \mathcal{T} \cup \mathcal{A}$ comprises a set of assertions
% $\mathcal{A}$ called the ABox (the extensional knowledge of the graph) and a set
% of axioms $\mathcal{T}$ called the TBox (the intensional knowledge of the
% graph). The axioms are expressed in a language $\mathcal{L}$ that is a fragment
% of first-order logic. 
% KGs can be seen as a special case of Knowledge Bases (KBs) where extensional knowledge is represented in the form of a graph.
% Regarding the structure of a KG, we adopt the following
% terminology with two main components: 
% % \begin{enumerate*}[label=\alph*)]
%   % \item 
%  {\em (i)} the terminological box (TBox), including classes, relations, rules, and
% restrictions, and 
% \todo{----->needed?}
%   % \item 
%   {\em (ii)} the assertional component (ABox), that describes
% specific instances based on the TBox definitions~\cite{Hogan2022}. 

% \todo[inline]{this is a good ref for here, but deals with KG instead of KB}
% \end{enumerate*}

% -Problem statement: Inconsistent KGs. Here we may add specific examples.
KGs are often developed through (semi-) automated knowledge extraction processes, and/or by integrating knowledge from different resources, which often lead to errors
and inconsistencies~\cite{Paulheim2016}. 
%We use the term ``error'' for a part of
%a KG that does not correspond to the real world and the term ``inconsistency''
%for a special case of error,
Inconsistencies, in particular, arise when parts of the TBox and/or ABox contradict.
% For example, there may be a TBox axiom stating that each person should be associated with a single city of birth. In practice, however, two different cities may be extracted as the birthplace of the same person. This could be the result of a named entity recognition where Athens, Greece is confused with Athens, Georgia. 
As inconsistencies hinder the process of reasoning, dealing with them is an important problem. 
We provide an overview of the state-of-the-art of reasoning on inconsistent KGs,
organized into three subtasks:
\begin{enumerate*}[label=\alph*)]
\item the detection of inconsistencies,
\item the fixing of inconsistencies, and
\item reasoning in the presence of inconsistencies. 
\end{enumerate*}
% In this direction, KGs can also be considered as a special case of Knowledge
% Bases (KBs) where knowledge is represented in the form of graphs. 

% -Scope of survey & difference from related surveys.
The scope of this survey is an overview of works that originate from different domains such as the Semantic Web, data quality, and query answering, highlighting why and how they are relevant to formal reasoning on inconsistent KGs. 
\citeauthory{Xue2022} provide a survey on the broader field of quality management in KGs and \citeauthory{Paulheim2016} provides a survey of approaches that infer missing knowledge (KG completion) or identify erroneous assertions (error detection).
In general, many works that focus on KG completion and error detection aim at improving the KG quality and some of them employ reasoning to add missing information or to detect errors. However, quite often they do not lead to a fully complete, error-free, and globally consistent KG. 
Conversely, our focus here is on formal reasoning in inconsistent KGs. In this context, some of the above approaches can be proven useful to move towards a consistent KG, for example just by removing the errors that exist; but they are not necessarily enough to achieve consistency, where inconsistency detection and fixing methods come into play. 
\citeauthory{Lambrix2023} provide a recent survey on the debugging and completion of ontologies, focusing on the task of repairing the TBox of a KB. We, on the other hand, pay more focus on repairing the ABox, which is much larger and error-prone than TBox for several real-world KGs, such as Bpedia~\shortcite{Auer2007} and YAGO~\shortcite{PellissierTanon2020}.
% , as explained in Section~\ref{sec:inconsistent_KGs}.  
% Several classes of methods are available for handling the lack of veracity in Big Data applications as surveyed by \citeauthory{Alevizos2018} and 
% \citeauthory{Chu2016} provide an overview of methods proposed for data repairing.  
% \citeauthory{Bell2007} fixing and tolerant. not user input, not KGs
% \citeauthory{Lambrix2023} ontologies as Tbox
In addition, for cases when KG consistency is not achievable, we also consider inconsistency-tolerant reasoning approaches~\cite{bienvenu_short_2020}. 
Alternative types of non-classical KG reasoning have been proposed, such as reasoning based on distributed representations and Neural Networks~\cite{Hohenecker2020,Chen2020}, which do not necessarily require consistency and lie beyond the scope of this survey.
% as our focus is on how to achieve formal logical reasoning for inconsistent KGs, accompanied by respective theoretical proofs and guarantees.

% \todo[inline]{+Crossover among/across fields.}
% \todo[inline]{ABox, TBox, and inconsistency ``types''}
% \todo[inline]{Running example: Find and introduce a running example to be used for all sections.}
% For example, a simple inconsistency can occur when a single entity is assigned two classes that are defined as disjoint in the TBox.
% \todo[inline]{Structure of this survey}
The remainder of the article is organized as follows:
First, we introduce some basic notions and terms about inconsistent KGs and
provide some examples in Section~\ref{sec:inconsistent_KGs}. Then we focus on
checking whether a KG is consistent or not, and what parts of it are involved in an inconsistency. This is a challenging task, in particular for large-scale KG, as we discuss in Section~\ref{sec:Check}. 
In Section~\ref{sec:Fixing} we investigate the fixing of an inconsistent KG, by altering its inconsistent parts, aiming for a consistent version on which reasoning is applicable. However, altering the KG to achieve consistency may not always be possible. 
In this direction, in Section~\ref{sec:Paraconsistent} we consider paraconsistent approaches that focus on tolerating the presence of inconsistencies during the reasoning process. 
Finally, Section~\ref{sec:Conclusions} presents a summary as well as the open challenges.

% Chen \textit{et al.}~\citeyear{Chen2020} explored different types of ``Knowledge reasoning'' on KGs, based on logic rules, distributed representation, and Neural Networks.
% \todo[inline]{Huang2005 is a good ref for the two alternatives fix vs non-fix.}
% \todo[inline]{table, approaches/citations.}

\section{Inconsistent Knowledge Graphs}
\label{sec:inconsistent_KGs}

A \gls{kg} $\mathcal{K} = \mathcal{T} \cup \mathcal{A}$ comprises a set of assertions
$\mathcal{A}$ called the ABox which expresses the extensional knowledge of the graph, and a set
of axioms $\mathcal{T}$ called the TBox which is the intensional knowledge of the
graph. An assertion is a triple $(s_1,l,s_2)$ representing a labeled edge of the \gls{kg}
between nodes $s_1$ and $s_2$ with label $l$. We will say that $s_1$ is the \emph{subject}, 
$s_2$ is the \emph{object} and $l$ is the \emph{predicate} of the triple.
The TBox is expressed in a language $\mathcal{L}$ that is a fragment
of first-order logic. In particular, it is usual to express the TBox in a \gls{DL}.
\glspl{DL} is a family of logics that are less expressive than 
first-order logic and are used to model \emph{concepts} (or classes), \emph{roles} (or relations) 
and their relationships. 

Concepts are used to model sets of nodes of the knowledge graph and roles model sets of edges and therefore correspond to unary and binary predicates 
of first-order logic, respectively. In \glspl{DL}, axioms are logical statements relating 
concepts and roles. The form of the axioms depends on the specific \gls{DL}, 
and as a result different \glspl{DL} have different expressive power.
The most basic language is $\mathcal{AL}$ (Attributive Language), and can be expanded \eg with $\mathcal{C}$ for complements to form $\mathcal{ALC}$, or $\mathcal{SR}$ for property chains, characteristics, and role hierarchies, $\mathcal{O}$ for nominals, $\mathcal{I}$ for inverse properties, and $\mathcal{Q}$ for qualified cardinality constraints to form the more expressive $\mathcal{SROIQ}$ language~\cite{baader2003description}.

\begin{example}
\label{exampleMain}
Consider the knowledge graph depicted in Figure~\ref{fig:KG_example}. The graph
consists of nodes like \texttt{Bob} and \texttt{Robby}. An edge labeled
\texttt{belongsTo} connects \texttt{Bob} to \texttt{Robby}. Similarly, an edge
labeled \texttt{hasName} connects \texttt{Robby} and \texttt{2000}. The nodes
\texttt{Person} and \texttt{Robot} also appear in the graph and represent
classes. The node \texttt{Bob} is linked to \texttt{Person}, indicating that
\texttt{Bob} is an instance of \texttt{Person}.

The TBox of the KG defines relationships between the classes. For example, the
axiom $\texttt{Person} \sqcap \texttt{Equipment} \sqsubseteq \bot$ expresses
that \texttt{Person} and \texttt{Equipment} are disjoint classes, and the axiom
$\texttt{Robot} \sqsubseteq \texttt{Equipment}$ states that \texttt{Robot} is a
subclass of \texttt{Equipment}. Additionally, the relation \texttt{belongsTo}
has domain and range restrictions for the classes \texttt{Equipment} and
\texttt{Person}, respectively. These restrictions are expressed by the axioms
$\texttt{Person} \equiv \forall \texttt{belongTo}.\top$ and $\texttt{Equipment}
\equiv \forall \texttt{belongTo}^{-1}.\top$. Data such as \texttt{2000} are treated 
as nodes in the knowledge graph. Data types are usually 
defined as classes that contain the corresponding values. For example, the node 
\texttt{2000} is an instance of \texttt{Integer} and \texttt{Integer} is disjoint with 
\texttt{String}.
\begin{figure}
\begin{center}
      \includegraphics[width=0.75\linewidth]{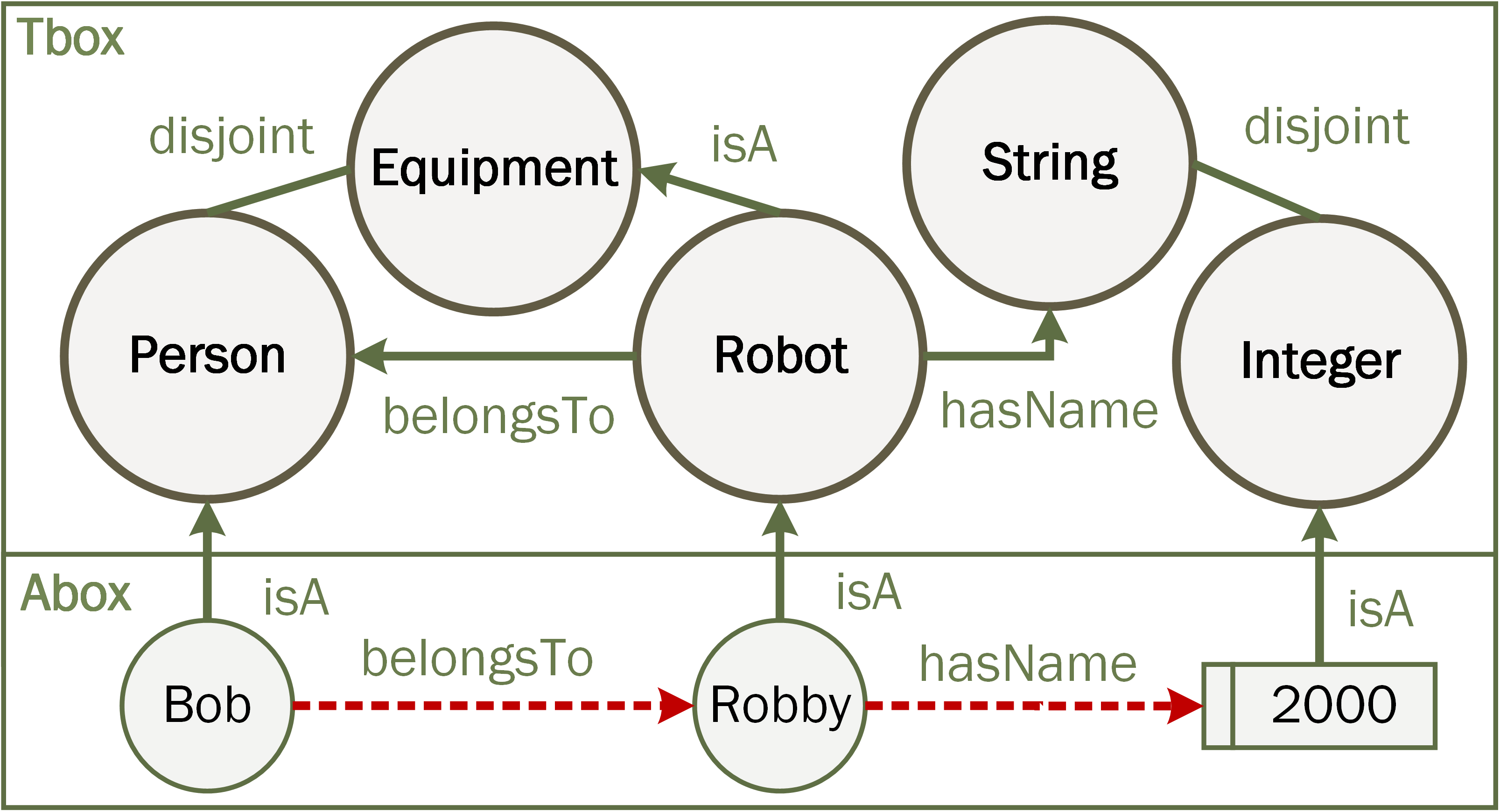}
      \caption{A KG example. Circles are classes and individuals in the TBox and
      ABox respectively. Rectangles are literal values. Arrows indicate properties
      (\texttt{belongsTo}, \texttt{hasName}) or inclusion (\texttt{isA}). Dashed
      arrows indicate ABox errors leading to inconsistency.}
      \label{fig:KG_example}
\end{center}
\end{figure}
\end{example}

%\paragraph{Reasoning}
A crucial feature of a \gls{kg} is the ability to reason about the knowledge it represents.
We consider the usual definition of an interpretation of first-order logical statements. 
Given a domain of individuals $\mathcal{D}$, an interpretation assigns a meaning to the non-logical symbols of the \gls{kg}.
For example, in the case of a \gls{kg} expressed in a \gls{DL}, 
an interpretation assigns to each concept $C$ a subset $C^\mathcal{D}$ of $\mathcal{D}$ 
and to each role $R$ a subset $R^\mathcal{D}$ of $\mathcal{D} \times \mathcal{D}$.
An interpretation is a \emph{model} of $\mathcal{K}$ 
if and only if it satisfies all axioms and assertions in $\mathcal{K}$. 
If $\mathcal{K}$ admits a model we will say that $\mathcal{K}$ 
is \emph{satisfiable}. We will say that $\mathcal{K}$ is \emph{unsatisfiable} 
if and only if no model exists. We also use the term \emph{consistent}
if and only if $\mathcal{K}$ is satisfiable and the term \emph{inconsistent} otherwise.
% We say that $\mathcal{K}$ \emph{entails} a formula $\phi$ if and only if
% every model of $\mathcal{K}$ is also a model of $\phi$ 
% and we write $\mathcal{K} \models \phi$.
%
\begin{definition}[Entailment]\label{def:entailment}
A KG $\mathcal{K}$ {\em entails} a formula $\phi$ ($\mathcal{K} \models \phi$), 
if and only if every model of $\mathcal{K}$ is also a model of $\phi$.
\end{definition}
If a \gls{kg} is inconsistent then any formula is entailed. 
In other words, we cannot distinguish between true and false 
assertions, and therefore reasoning is useless.

\begin{example}
Consider again the \gls{kg} in Figure~\ref{fig:KG_example}
and assume that due to some error in knowledge extraction, an
assertion is generated stating that \texttt{Bob}
\texttt{belongsTo} \texttt{Robby}.
This leads to the following inconsistencies:
\begin{enumerate}[label=(\roman*)]
\item $\texttt{Bob} \in \texttt{Person}$, while the domain of \texttt{belongsTo} 
      should be \texttt{Robot} and \texttt{Robot} is disjoint with \texttt{Person};
\item $\texttt{Robby} \in \texttt{Robot}$ and $ \texttt{Robot} \sqsubseteq \texttt{Equipment}$, 
      which is disjoint with \texttt{Person} that is defined as the range of
      \texttt{belongsTo}.
\end{enumerate}
Moreover, another TBox axiom may restrict a data property \texttt{hasName} to
hold only values of type \texttt{xsd:string}. If the ABox contained, \eg that
\texttt{Robby} \texttt{hasName} \texttt{"2000"\string^\string^xsd:int}, then
this would be the cause of an additional inconsistency, since
\texttt{xsd:string} and \texttt{xsd:int} are different data types.
\end{example}

There can be several reasons for an inconsistent \gls{kg}.
The TBox may be problematic regardless of the ABox. 
In this case, the set of axioms are ill-formed and produce contradictory conclusions. 
Furthermore, the ABox may be inconsistent regardless of the TBox.
Finally, both ABox and TBox may be consistent by themselves but inconsistent 
if combined.
% This, e.g. may happen if a class is defined to be simultaneously a subclass of two disjoint classes,; 
% \item an individual is asserted (or inferred) to either belong to disjoint classes (e.g. $a_1 \in A^{\mathcal{I}}$, $a_1 \in B^{\mathcal{I}}$, and $A^{\mathcal{I}} \cap B^{\mathcal{I}} = \emptyset$), or to violate property domain or range restrictions (e.g.  $(a_1, a_2) : R$, $a_1,a_2 \in A^{\mathcal{I}}$, $\forall R.B^{\mathcal{I}}$, and $A^{\mathcal{I}} \cap B^{\mathcal{I}}=\emptyset$) ;
%Apart from referring to individuals instead of classes, the difference between this and the previous case is that here a DL reasoner cannot conclude anything meaningful. 
% and 
% \item a data type is asserted which contradicts the domain or range defined by an existing TBox axiom (e.g. for individual $a_1$ and literal value $v$ and data types $DT_1$, $DT_2$, then $a_1 : v\in DT_1$, $a_1 \in A^{\mathcal{I}}$, and $\forall A^{\mathcal{I}} \in DT_2$ )~\cite{schlobach2003non,Topper2012,lertvittayakumjorn2017resolving}.
% In terms of logics, an inconsistent ontology is one that does not have a model.
%
Inconsistencies may be a result of ill-defined TBoxes, 
or when definitions from multiple sources are mixed,
\eg in cases of knowledge migration or KG merging~\cite{huang2005reasoning}.
% A consistent support set for $\phi$ w.r.t. $\mathcal{K}$ is a subset $S \subseteq \mathcal{K}$
% such that $S$ is consistent and $S \models \phi$. If $S$ is minimal, i.e., there is no 
% $S' \subset S$ such that $S' \models \phi$ then $S$ is a \emph{justification} for $\phi$.
\begin{definition}[Conflict set, Explanation]\label{def:conflictset}
A \emph{conflict set} is a subset $S \subseteq \mathcal{K}$ such that $S$ is inconsistent.
If $S$ is minimal then $S$ is an inconsistency \emph{explanation}. 
\end{definition}

In order to re-establish consistency in a \gls{kg} we need to 
alter the TBox, the ABox or both. In general, we can define a distance function $\Delta$
over pairs of KGs and a partially order set $(Q, \preceq)$ that quantifies the degree of alteration among different KGs. 
Given two KGs $\mathcal{K}$ and $\mathcal{K}'$, $\Delta(\mathcal{K}, \mathcal{K}')$ maps to an element 
of $Q$. Given KGs $\mathcal{K}$, $\mathcal{K}_1$ and $\mathcal{K}_2$ we will say that $\mathcal{K}_1$
is at least as close as $\mathcal{K}_2$ is to $\mathcal{K}$, 
if and only if $\Delta(\mathcal{K}, \mathcal{K}_1) \preceq \Delta(\mathcal{K}, \mathcal{K}_2)$. Also, we will say that 
$\mathcal{K}_1$ is $\Delta$-minimally altered from $\mathcal{K}$ if and only if there is no \gls{kg}
$\mathcal{K}'$ such that $\Delta(\mathcal{K}, \mathcal{K}') \preceq \Delta(\mathcal{K}, \mathcal{K}_1)$.
\begin{definition}[Repair]\label{def:repair}
Let $\Delta$ be a distance function between KGs and $\mathcal{K}$ be a \gls{kg}.
A $\Delta$-\emph{repair} $\mathcal{R}$ is a $\Delta$-minimally altered version of $\mathcal{K}$ that is consistent. 
\end{definition}
Several alterations can occur such as 
to insert or to delete assertions and axioms of the original KG. There are also various 
ways to define $\Delta$, where one example is to consider the number of changes.
A common case is to consider only deletions and define
$\Delta(\mathcal{K}, \mathcal{R}) = \mathcal{K} \backslash \mathcal{R}$ if $\mathcal{R} \subseteq \mathcal{K}$ 
and $\Delta(\mathcal{K}, \mathcal{R}) = \mathcal{R}$ otherwise.
In this case, a repair $\mathcal{R}$ is a $\subseteq$-maximal consistent subset $\mathcal{R} \subseteq \mathcal{K}$, that is, 
any superset $\mathcal{K} \supseteq \mathcal{R}' \supset \mathcal{R}$ is inconsistent. 
It is also a common case to focus on repairs that remove only assertions from the ABox and retain all axioms of the TBox.
Those repairs are referred to the literature as \glspl{AR}.
In general, we can obtain a repair by removing the least knowledge (i.e. axioms or assertions) required, so that no conflict set is contained in the \gls{kg} after the removals.
% Apparently, a repair should not contain any conflict set.

% \noindent{\bf TBox:}

% {\small

% \noindent\texttt{SubClassOf(Robot Equipment)}

% \noindent\texttt{ObjectPropRange(belongsTo Person)}

% \noindent\texttt{ObjectPropDomain(belongsTo Equipment)}

% \noindent\texttt{DataPropRange(hasName str)}

% \noindent\texttt{DisjointClasses(Equipment Person)}
% }

% \vspace{+4pt}
% \noindent{\bf ABox:}

% {\small 
% \noindent\texttt{ClassAssertion(Person Bob)}

% \noindent\texttt{ClassAssertion(Robot Robby)}

% \noindent\texttt{ObjectPropAssertion(belongsTo Bob Robby)}

% \noindent\texttt{DataPropAssertion(hasName Bob "BOB":str)}

% \noindent\texttt{DataPropAssertion(hasName Robby "123":int)}
% }

% \vspace{+4pt}
% \noindent{\bf Inconsistency Explanations:}

% {\small
% \noindent\texttt{DataPropAssertion(hasName Robby "123":int)}

% \noindent\texttt{DataPropRange(hasName str)}
% \vspace{+4pt}

% \noindent\texttt{ClassAssertion(Robot Robby)}

% \noindent\texttt{SubClassOf(Robot Equipment)}
% \noindent\texttt{ObjectPropDomain(belongsTo Person)}
% \noindent\texttt{ObjectPropAssertion(belongsTo Robby Bob) }
% \noindent\texttt{DisjointClasses(Equipment Person)}

% \vspace{+4pt}

% \noindent\texttt{ClassAssertion(Person Bob) }

% \noindent\texttt{ObjectPropRange(belongsTo Equipment) }

% \noindent\texttt{ObjectPropAssertion(belongsTo Robby Bob)}

% \noindent\texttt{DisjointClasses(Equipment Person) }

% }

% \subsection{Problems and tasks}

% The key problem is that i
If a \gls{kg} is inconsistent then reasoning 
% in the classical sense, 
is uninformative. In other words, even an insignificant 
piece of contradictory information will render other entailments inaccessible even though 
their validity may not depend on this information. In that sense, effective ways for dealing with 
inconsistent KGs are crucial.

Even deciding whether a \gls{kg} is inconsistent can be a hard task. 
This decision problem is also termed as satisfiability decision and for 
expressive languages, such as \ALC{}, it lies within the \PSPACE complexity class~\cite{Donini_2007}.
A second task related to consistency checking is the equally hard task 
of computing the inconsistency explanations, \ie the minimal sets that cause an 
inconsistency in the \gls{kg}. 

% \todo[inline]{say something about measuring how inconsistent is a graph}
% KG fixing is the task where given an inconsistent KG we come up with a new version of it that is consistent, by altering particular parts that are problematic. This involves selecting an appropriate repair and updating the KG accordingly to resolve all the conflicts that exist. Therefore, the main requirement in the KG-fixing task is the consistency of the KG. In parallel, as there may be a vast amount of alternative sets of update actions that can make a KG consistent, calculating these sets and choosing among them can be very demanding in computational terms, hence the efficiency of the fixing process is also required. Finally, although the main aim of KG-fixing is not to remove all the errors or add all the knowledge missing from a KG, repairs with fewer errors and more complete repairs can be preferred, even if \eg repairs with fewer update actions are available. 
% If there is no desire to repair the KG, the alternative 
% is to reason in a non-standard way even in the presence of inconsistent data. 
% According to this paradigm, a new logical consequence relation is defined such that reasoning is not
% trivial, that is, not every formula ends up being entailed by the KG.

% We now proceed to describe key existing approaches for each of these tasks.
In the section that follows, we survey the approaches on KG inconsistency
detection. Subsequently, we review the literature on handling such
inconsistencies. 

\section{Inconsistency Detection}\label{sec:Check}
% \todo[inline]{Charis}

% \subsection{Detection of inconsistency}
% \todo[inline]{for each, map to particular inconsistency types. Classify according to exact or approximate. Highlight criteria. Objectives and motivation.}

To detect inconsistencies, i.e. to come up with the inconsistency explanations (cf. Def.~\ref{def:conflictset}), we could rely on exact or approximate
approaches. 
Exact methods would generate a set of explanations, which \eg in the case of the example of Figure~\ref{fig:KG_example}, would be the following:
\begin{namedtheorem}
\label{expExample}
(1) \texttt{Equipment DisjointWith Person},
(2) \texttt{belongsTo Domain Equipment},
(3) \texttt{Bob belongsTo Robby},
(4) \texttt{Bob Type Person}.
\end{namedtheorem}
\begin{namedtheorem}
\label{expExample2}
(1) \texttt{Equipment DisjointWith Person},
(2) \texttt{Robot SubClassOf Equipment},
(3) \texttt{belongsTo Range Person},
(4) \texttt{Robby Type Robot},
(5) \texttt{Bob belongsTo Robby}.
\end{namedtheorem}
\begin{namedtheorem}
\label{expExample3}
(1) \texttt{hasName Range: xsd:string}\\
(2) \texttt{Robby hasName "2000"\string^\string^xsd:int}.
\end{namedtheorem}
\noindent
% Following Definition~\ref{def:conflictset}, the KG conflict sets $S$ would consist of all triples that are included in an explanation. In our example, 
Consider the case of Explanation~\ref{expExample} where the size of the conflict set $S$ is four.
If we removed any of the triples, resulting to an $S'\subseteq S$, then no inconsistency would be present.
If we removed (1), due to the open world assumption~\cite{baader2003description} under which most DL reasoners operate, there would be no conflict with \texttt{Bob} being a \texttt{Person}; if we removed (2), then there would be no restriction for (3), and so on.
% Thus,~\ref{expExample},~\ref{expExample2}, and~\ref{expExample3} are explanations.

Table~\ref{tab:detect} outlines the most relevant methods for KG inconsistency detection, which we now describe.
\begin{table}
\begin{center}
\begin{tabular}{clll}
\toprule
 & Reference (Implementation)   & Language & Complexity \\
\hline
\multirow{4}{*}{\rotatebox[origin=c]{90}{Exact}}
% \scriptsize {\tiny \cite{haase2005framework}} {\bf *}   & Exact & $\mathcal{SHOIN}(\mathbf{D})$ & \textsc{NExpTime} \\
& {\shortciteA{HorridgeExplanations2009}} \checkmark &  $\mathcal{SROIQ}$ & \textsc{2NExpTime} \\
& {\shortciteA{suntisrivaraporn2008modularization}} {\bf *}, $\mathbf{\parallel}$ & $\mathcal{SHOIQ}$ & \textsc{NExpTime} \\
&  {\shortciteA{tran2020fast}  $\mathbf{\parallel}$} &  $\DLLite_\mathit{bool}^\mathcal{H,+}$ & \textsc{NPTime}  \\
& {\shortciteA{meilicke2017fast}} {\bf +} & $\supseteq$\DLLite$_\mathcal{A}$ & \textsc{PTime} \\
 % $\subset\mathcal{SROIQ}(\mathbf{D})$  &  \textsc{NExpTime} ($\mathbf{\parallel}$)\\
\hline
% \scriptsize {\tiny \cite{roussey2013antipattern}} & Approx. & ? & ?\\
\multirow{5}{*}{\rotatebox[origin=c]{90}{Approximate}}
&  {\shortciteA{de2021analysing}} \checkmark, $\mathbf{\parallel}$  & $\mathcal{SHOIN}$ & \textsc{NExpTime} \\
&  {\shortciteA{paulheim2016fast}} $\mathbf{\parallel}$ & $\mathcal{SHIN}(\mathbf{D})$ & \textsc{PTime} \\
&  {\shortciteA{Paulheim2014} \checkmark}  & $\mathcal{SROIQ}(\mathbf{D})$  & \textsc{PTime} \\ %\# triples$^2$\\
% \scriptsize {\tiny \cite{hong2023inconsistency}} & Approx. & $\mathcal{SROIQ}(\mathbf{D})$ & ?\\
&  {\shortciteA{10.1145/3543873.3587536}} \checkmark, $\mathbf{\parallel}$  & $\mathcal{SROIQ}(\mathbf{D})$ & \textsc{PTime}\\
&  {\shortciteA{chen2021neural}}  & $\mathcal{SROIQ}(\mathbf{D})$  & \textsc{PTime} \\
% &  {\shortciteA{shchekotykhin2014sequential} \checkmark} &  $\mathcal{SROIQ}$  &  \textsc{LogTime} \\
\bottomrule
\end{tabular}
\caption{Implementation availability, language expressivity and complexity of approaches for inconsistency detection. `{\tiny \checkmark}' ~indicates a publicly available implementation, `{\bf *}' initially published but currently unavailable, and `{\bf +}' available on request. `$\mathbf{\parallel}$' indicates opportunity for parallel execution. Complexity is in terms of the size of the ABox. For detailed descriptions of the symbols that define the language fragment support, we refer the reader  to~\citeauthory{baader2003description}. 
}
\label{tab:detect}
\end{center}
\end{table}
\citeauthory{HorridgeExplanations2009} propose sound and complete methods 
% to calculate inconsistency explanations is to
which exhaustively search for explanations by recursively employing the hitting set tree algorithm~\cite{reiter1987theory} on subsets of the initial KG.
%incrementally add or remove triples until 
%an explanation
% a minimal subset that includes an
% inconsistency 
%remains~\cite{HorridgeExplanations2009}, indicating the presence of inconsistency. 
% For example, \citeauthory{corcho2009pattern} detects unsatisfiable classes and can be used to guide less DL-savvy domain experts based on design antipatterns to correct the inconsistencies.
Unfortunately, such
approaches do not scale to large KGs. 
% An answer to the scalability issues is to adopt 
To address this issue, modularization techniques may be adopted.
Modularization has been introduced in the scope of reusing and extending
ontologies, and aim to maintain only the triples that are effectively used for the inference of new knowledge, this way reducing the size of
the utilized KG. 
For example, \citeauthory{suntisrivaraporn2008modularization} extract locality-based modules
and employ hitting set trees to discover explanations of \SHOIQ{}
entailments, though this approach does not focus solely in detecting inconsistencies. 
% Although not focusing on inconsistencies, this approach may be
% applied for detecting erroneous triples as well.
% In~\cite{fahad2012detection} multiple inconsistency detectors are employed for the task of merging different ontologies. 
% Based on mappings generation between the ontologies to be merged, conflicts in generalized concept inclusions are discoverd along with disjoint relations.

\citeauthory{tran2020fast} decompose KGs into modules of individuals, \ie
subsets of triples of the original KG that include a reference to a particular
individual.
Each module along with the TBox---generally being of significantly
smaller size than the original large KG---is analyzed independently and in
parallel to obtain the inconsistency explanations. 
Considering the example of Figure~\ref{fig:KG_example}, the module of \texttt{Bob} would include the TBox and two ABox assertions, \texttt{Bob} $\in$ \texttt{Person} and \texttt{Bob belongsTo Robby}, thus allowing the detection of the inconsistency of Explanation~\ref{expExample}.
Similarly, the module of \texttt{Robby} will contain all ABox assertions apart from \texttt{Bob} $\in$ \texttt{Person}, thus allowing the detection of Explanations~\ref{expExample2} and~\ref{expExample3}.
\citeauthor{tran2020fast} employ an additional step that groups explanations into abstractions,
which describe high-level types of inconsistency. This way, we may present to the human KG developer a smaller number of
abstractions, instead of a larger number of
explanations that may be overwhelming. Although scalable, this
approach works on $\DLLite^\mathcal{H,+}_\mathit{bool}$, which is only a fragment of OWL 2~\cite{dllite}.
For instance in Example~\ref{exampleMain}, if instead of the \texttt{Bob belongsTo Robby} assertion we had \texttt{Bobby belongsTo Robby} and \texttt{Bob sameAs Bobby}, then Explanation~\ref{expExample} would also include \texttt{(5) Bob sameAs Bobby}, but would nevertheless not be detected at any module of the corresponding individuals by~\citeauthory{tran2020fast}; the module of \texttt{Bob} would not contain \texttt{Bobby belongsTo Robby}, and the module of \texttt{Bobby} would not contain \texttt{Bob Type Person}.
%, and in particular the
%existential quantifier 
%\texttt{ObjectSomeValuesFrom(Object Property Expression, Class Expression)} 
%supports only the \texttt{owl:Thing} class expression, 
%thus it is not possible to detect inconsistencies of particular types, e.g. those inflicted by incompatible (wrt. the TBox)
%chain property relations. 
% It is also worth noting that the source code that this work relies upon is not openly available.

Another way to achieve tractability is to reduce the expressiveness of the
supported language. Towards this, \citeauthory{meilicke2017fast} utilize clash
queries for \DLLite$_\mathcal{A}$ that allow the detection of inconsistencies that lie beyond the expressivity restrictions of this language for a given TBox and a sequence of different ABoxes.
Clash queries are pre-compiled combinations of classes, relationships and attributes that are capable of inducing inconsistency in \DLLite$_\mathcal{A}$ KGs.
Consider Example~\ref{exampleMain} again, on which we could create an appropriate query to retrieve the triples that satisfy $\mathcal{T} \models \rho(U) \sqsubseteq T$ and $U(\alpha,v) \in \mathcal{A}$ and $v^{\mathcal{I}} \notin T^{\mathcal{I}}$, where $\rho(U)$ is an attribute range of values, $T$ a value domain, $U(\alpha,v)$ the assignment of an attribute value $v$ for an individual $\alpha$, and $v^{\mathcal{I}}$ and $T^{\mathcal{I}}$ interpretation functions.
Then, the result would be Explanation~\ref{expExample3}.
Different types of inconsistencies can be captured by other clash patterns, similarly to the above.
% the following pattern: . If we submitted a query for the pattern of a value domain of a role, for which an ABox assertion assigns this kind of value as an attribute to an individual
% , \ie pre-compiled combinations of classes, relationships, and attributes that allow us to detect inconsistencies given a TBox, and a sequence of ABoxes that lie beyond the expressivity restrictions of \DLLite$_\mathcal{A}$.
% $\DLLite_{\mathcal{A}}$. 
Likewise in~\citeauthory{meilicke2017fast}, with the help of a classical reasoner, the
signature elements that are entailed by a clash pattern are matched to those
that induce inconsistency.\footnote{The signature corresponds to the set of entities that are included in an expression or formula.} This task is performed incrementally as more ABoxes
are examined, and for signature element combinations that have already been
recognized as inconsistent, reasoning is skipped. Thus, the
amount of computationally expensive invocations of the reasoner decreases over
time.
%However, the detection of all possible inconsistency types in more expressive languages is not guaranteed 
%as some lie beyond the expressivity capabilities of clash queries.

The other category of approaches that achieve scalability generally sacrifice completeness, and
thus rely on approximate reasoning. For instance, an approximate approach that is similar to~\citeauthory{tran2020fast} that we already described, is that of
\citeauthory{de2021analysing}.
% follows similar principles
% as~\cite{tran2020fast} for grouping inconsistency types and presenting them to a human experts for further corrective actions, but splits the initial KG into subgraphs in a different, heuristic,
% way. 
The difference, however, is that instead of starting from the individuals to extract their modules, this work extracts subgraphs, based on all
 subjects that are included in the ABox assertions. Each such subject constitutes the root node
for the construction of each respective subgraph. Then, according to breadth-first search, all the triples with a
subject that corresponds to the root node are included. Next, the objects and
predicates are expanded accordingly, up to a maximum number $G_{\mathit{max}}$ defined
by the user. 
$G_{\mathit{max}}$ tunes
the tradeoff between completeness and scalability.
Then, for each subgraph an off-the-shelf reasoner is invoked to
obtain inconsistency explanations, which are finally grouped into the so-called
``anti-patterns'' by substituting subjects, objects and, in some cases, predicates
as well, with variables. 
Intuitively, an anti-pattern is a smaller, inconsistent subgraph with variables instead of values that serves as an abstraction template for categorizing the inconsistency types that are discovered in the KG.
The concept of $G_{\mathit{max}}$ could also be incorporated in the approach of~\citeauthory{tran2020fast} for combining modules of individuals up to a number of `hops', and thus cover a more expressive language fragment. 
Another difference with that work is that the root node of the subgraph can be any entity, instead of merely individuals.

% It is worth noting that no formal guarantees are provided, as this approach is only evaluated empirically.

% In~\cite{roussey2013antipattern}, the KG is
% transformed and potentially unsatisfiable classes are detected by 
% SPARQL queries. These are in turn presented to a human evaluator that either
% confirms or rejects the inconsistency warning. Although this approach does not
% require the use of a DL reasoner as it avoids to deal with class disjointness,
% it may only serve as an early warning method instead of an accurate
% inconsistency detecting one.

\citeauthory{paulheim2016fast} present an approximate method for
predicting if a set of ABoxes are inconsistent with respect to a TBox. This is performed by formulating the
problem as binary classification, \ie trying to determine whether a given ABox is consistent or not, judging by a set of feature values.
% This is a process that is expected to be faster than invoking a reasoner. 
The ABoxes are translated into feature vectors
using path kernels, and then off-the-shelf classification algorithms are
applied. 
According to the empirical analysis, this method can scale even for very large KGs. Moreover, key factors affecting performance are
 the feature translation and classification, while accuracy is shown to reach up to 96\% of correct class label predictions for some dataset cases. 
\citeauthory{Paulheim2014} consider the statistical distributions of types in the subject and object position linked with a property in an assertion, to calculate a confidence score and highlight those surpassing a given threshold as erroneous.

\citeauthory{10.1145/3543873.3587536} proposes an approximate method for
detecting anomalous---and thus, possibly erroneous---assertions in KGs, which relies on unsupervised learning for anomaly detection. Following the Corroborative Path algorithm that is a variant of the Path Rank algorithm~\cite{10.1145/1835804.1835916}, this method identifies alternative paths between entities in a triple, \ie the existence or not of other property relations (or chains of such) between the subject and the object. 
This way, the method constructs a binary feature matrix with the entities as rows, and the properties (or property chains) as columns.
 The KG entities are also assessed regarding their correctness by considering other kinds of features, such as graph structure characteristics, frequency of appearance, etc. Finally, a one-class support vector machine identifies the anomalous parts, which subsequently should be removed or corrected.

Other approaches rely on neural networks as
in~\citeauthory{chen2021neural}. TBox axioms and ABox assertions are transformed to embeddings, namely vectors with numerical values that result from TransE~\cite{NIPS2013_1cecc7a7} transformations of the original KG data. The values of the vector coefficients are such that their distance indicates the degree of relationship between the original data~\cite{Ji2022}.
After the KG transformation to embeddings, a fitness score
is assigned to each element according to its degree of expectancy according to the transformation mechanism.
Also, additional attention coefficients are taken into account \ie according to if the subject
satisfies different types of axioms, such as domain, range, disjointness, irreflexive, and asymmetric  axioms.
All these scores are combined into a cross-entropy loss function which is minimized for training the neural network. 
However, such methods are not guaranteed to detect every possible inconsistency in a KG.
% Although fast, such methods may face issues in cases where erroneous assertions come in large numbers.

There are also relevant approaches from other domains, such the QuickXplain algorithm and its variants~\cite{junker2004preferred,shchekotykhin2014sequential} that seek to explain constraint violations in constraint satisfaction problems by dividing and pruning the initial input graphs.
However, since these mainly rely on setting preference orderings on the constraints---a procedure that is not straightforward for real-world large KGs---we consider them out of the scope of our survey.
Overall, there are multiple algorithms that can detect inconsistencies in KGs, which are based on different principles and aim to strike a balance between completeness and scalability.
Having analyzed their functionality, we now proceed with the next step on how to deal with reasoning over inconsistent KGs \ie to fix or tolerate the inconsistencies.
% \citeauthory{junker2004preferred} proposed QuickXplain, a generic approach for obtaining inconsistency explanations in constraint satisfaction settings. According to this, the problems are assessed in a divide and conquer manner, also allowing to define preference orderings for conflicts and relaxations. The work of~\citeauthory{shchekotykhin2014sequential} presents a QuickXplain modification that is applied for faulty knowledge base diagnosis. Employing such approaches for large KGs however, may not be straightforward, since, \eg the definition of preferences cannot be performed manually given the great number of axioms and assertions.

% \cite{gueffaz2012inconsistency} with model checking.
% \cite{fahad2012detection} for mapping based approaches.
% Finally, the more recent work of~\cite{hong2023inconsistency} focuses on the
% semantic information encapsulated in the paths of the KG. Multiple features are
% calculated, i.e., the semantic value of entity pairs, the entity-path
% similarity, and the relative path confidence, which are then fed into an
% off-the-shelf classifier. This approach is shown to outperform other
% well-known techniques such as TransE and CKRL.

%\newpage

\section{KG Fixing}\label{sec:Fixing}
% \todo[inline]{Lead:Tasos}
% We focus on the task of `fixing' an inconsistent KG, i.e. selecting an appropriate repair, as defined in Definition~\ref{def:repair}. 
% % That is, given an inconsistent KG, obtaining a new version of that KG that is consistent.
% KG fixing involves a set of actions that change the KG in order to resolve any inconsistency. We note such a set of actions leading to a repair as a ``fix'', as defined below. 
We focus on the task of `fixing' an inconsistent KG, i.e. selecting and applying a set of actions that change the KG in order to resolve any inconsistency and lead to a repair, as defined in Definition~\ref{def:repair}. 
% That is, given an inconsistent KG, obtaining a new version of that KG that is consistent.
% KG fixing involves a set of actions that change the KG in order to resolve any inconsistency. We note such a set of actions leading to a repair as a ``fix'', as defined below. 
\begin{definition}[fix]\label{def:fix}
% Let $\Delta$ be a distance function between KGs and $\mathcal{K}$ be a \gls{kg}.
A \emph{fix} $\mathcal{F}$ is a set of alterations on $\mathcal{K}$ that lead to a $\Delta$-\emph{repair} $\mathcal{R}$. 
\end{definition}

% To the best of our knowledge, no survey is available for this specific task of KG repairing. However, works relevant to this task have been considered in surveys of the related tasks of KG quality management and KG refinement.
In the example of Figure~\ref{fig:KG_example}, a potential fix could be, for instance, to update the ABox as presented in the fix of Example~\ref{fixExample}:
\begin{example}
\label{fixExample}
Add \texttt{Robby} \texttt{belongsTo} \texttt{Bob} in place of \texttt{Bob} \texttt{belongsTo} \texttt{Robby}, and add \texttt{Robby} \texttt{hasName} \texttt{"2000"\string^\string^xsd:string} in place of \texttt{Robby} \texttt{hasName} \texttt{"2000"\string^\string^xsd:int}.
\end{example}
% subject and object of the \texttt{BelongsTo} ABox assertion with \texttt{Robby} and \texttt{Bob} respectively, and change the value \texttt{"2000"\string^\string^xsd:int} of the \texttt{hasName} ABox assertion into \texttt{"2000"\string^\string^xsd:string} (\textit{repair example}). 
In addition to the above, there are alternative fixes that lead to a consistent KG, such as removing the assertions involved in the inconsistency from the ABox and/or the TBox. In the remaining part of this section, we provide an overview of how related works address or can be used to address the KG fixing task, categorized according to the part of the KG that is considered for updating.

\subsection{Unreliable TBox}
% TBox only
An inconsistency in a KG can be the result of an ill-defined TBox. 
For instance, adding that class \texttt{Robot} is a subclass of \texttt{Person} (\texttt{Robot} $\subseteq$ \texttt{Person}) in the TBox of the example of Figure~\ref{sec:inconsistent_KGs} would lead to an inconsistent KG. This is because the class \texttt{Robot} is also a subclass of \texttt{Equipment} which is disjoint with \texttt{Person}, hence it contradicts the assertion that \texttt{Robby} \texttt{isA} \texttt{robot}. 
% \todo{unsatisfiable ----class}
In such cases, fixing the KG should be achieved by changing the TBox to render it free of such contradictions. A reliable source for such resolution of inconsistencies is domain experts. However, humans need to be guided throughout the elements of the TBox that are problematic, given that the number of such elements is manageable. 
% and if many where to start from, as this can affect the whole process significantly. 

Several alternative methods are available in this direction, aiming to minimize the input required by the domain experts in order to resolve the faults in the ontology, make it more complete, or revise it~\shortcite{Lambrix2023,Qi2008}.
In this direction, some methods rank the ontology elements based on the number of inconsistencies they are involved in~\shortcite{Heyvaert2019}, belief-revision approaches considering the number of other axioms that can be automatically evaluated upon the evaluation of each axiom~\shortcite{Nikitina2012}, or considering parts of the ontology that can be trusted in combination with consequences of alternative update actions~\shortcite{Penaloza2019}.
Other methods develop query-based fault-localization strategies that rely on active learning~\shortcite{Rodler2016}, axiom likelihood estimation~\shortcite{Shchekotykhin2012}, or consider different query-answering behaviors by the experts~\shortcite{Rodler2022}.
Finally, a tool, named \textit{OntoDebug}~\cite{Schekotihin2018}, is also available for guiding experts to resolve errors in ontologies, as a plug-in for the popular open-source ontology editor Prot\'eg\'e~\cite{Musen2015}.
% ~\shortcite{Rodler2022,Heyvaert2019,Penaloza2019,Schekotihin2018,Rodler2016,Shchekotykhin2012,Nikitina2012}.
% For example, the \algo{resglass} method provides an order with which rules and ontology elements should be inspected (manually), based on the number of inconsistencies they are involved in \shortcite{Heyvaert2019}.
%  Query-based fault localization (QFL): Query answering strategies
% Rodler2016: Activle learning for interactive debugging.
% Shchekotykhin2012: entropy-based strategy that considering the likelihood of axioms in need of change.
% Rodler2022: we differentiate between seven different expert types with regard to their query answering behavior, e.g., whether queries as a whole or the individual query...
% Nikitina2012: axiom impact: the number of axioms that can be automatically evaluated upon approval or decline of an axiom.
% Penaloza2019: considering parts of the ontology that can be trusted in combination with consequences of alternative update actions. 

% \textbf{ABox and TBox together}:
In the more general case, inconsistency is caused by
% a combination of TBox and ABox axioms, as in the example of Figure~\ref{fig:KG_example}. In such cases, the 
errors in any of the axioms involved and the fix may require altering the TBox, the ABox, or both. For example, instead of updating the value \texttt{"2000"\string^\string^xsd:int} into \texttt{"2000"\string^\string^xsd:string} to comply with the TBox, as suggested in the fix of Example~\ref{fixExample}, one could remove the restriction itself. This depends on whether an integer should be acceptable as an equipment name or not.   
Apart from removing elements from the TBox, adding and/or altering elements may also be possible. A method towards this direction was proposed by \citeauthory{Topper2012}, where the TBox may be updated with new restrictions during the KG development process to detect as many inconsistencies as possible, identifying errors. During this process, altering the elements of both the ABox and the TBox is supported, to resolve the detected inconsistencies.
% \begin{itemize}
%     \item Manual repair: Suggest domain/range/disjointness constraints (TBox) or edges (ABox) to remove. (``...transform semantic errors into logical ones by extending the axioms of the underlying ontology to cause a logical contradiction that can then be recognized by a reasoner.''~\cite{Topper2012})
% \end{itemize}

\subsection{Reliable TBox}
Although errors in the TBox are possible, a lot of research focuses on inconsistencies caused by erroneous ABox assertions. In such cases, the TBox is considered reliable and the fix consists of updating the ABox, as performed in the fix of Example~\ref{fixExample}. In many KGs, such as  DBpedia~\shortcite{Auer2007} and YAGO~\shortcite{PellissierTanon2020}, this is a reasonable scenario as the ABox can often be very large rendering its manual development or validation less feasible than that of the TBox, which can be typically smaller. A discussion on this topic is provided by~\citeauthory{Paulheim2016}.
In the remainder of this section, we present some methods for fixing a KG with a reliable TBox. We focus on how alternative fixes are generated and chosen, and what types of constraints are considered, as summarized in Table~\ref{tab:fixing}.

\begin{table}
\begin{center}
\begin{tabular}{L{0.39\linewidth}L{0.35\linewidth}L{0.25\linewidth}}
% {lll}
\toprule
Reference (Implementation)   & Constrains considered                 & Fix selection \\
\midrule
\shortciteA{Arioua2018}                    & Relational database constraints 
% (Tuple-generating and conflict-detecting dependencies) 
% TGDs and CDDs
& User input            \\
\shortciteA{Fan2019}                       & Graph-quality rules          & User input, Reliability levels        \\
\shortciteA{Bonatti2011}                   & $O2R^-$ subset of OWL 2 RL & Reliability levels            \\
\shortciteA{Benferhat2015}                 & Prioritized $\DLLite$                & Reliability levels            \\
\shortciteA{Ahmetaj2022} \checkmark        & SHACL                              & Action minimality     \\
\shortciteA{Baader2022} \checkmark         & $\mathcal{EL}$                &  Entailment-loss minimality    \\
Pellissier Tanon et al. \citeyear{PellissierTanon2019,PellissierTanon2021}  & RDF correction rules               & User input, Supervised ML       \\

\shortciteA{Paulheim2014} \checkmark       & RDF properties and types                & Unsupervised ML           \\
\shortciteA{Cheng2020}                     & Graph Repairing Rules and Flexible Graph Repairing Rules      & Unsupervised ML           \\
\shortciteA{Melo2020} \checkmark           & SHACL                              & Unsupervised ML, External Knowledge       \\
\shortciteA{Abedini2022}                   & Unique RDF relations             & Unsupervised ML           \\
\shortciteA{Ye2023}                        & Weighted first-order-logic rules         & Unsupervised ML  \\
\shortciteA{Gad-Elrab2019} \checkmark      & Horn clauses                       & User input, External Knowledge         \\
\shortciteA{Arnaout2022}                   & OWL 2                               & External Knowledge            \\
\bottomrule
\end{tabular}
\caption{Overview of fixing approaches considering a reliable TBox. \checkmark indicates a publicly available implementation. ML stands for Machine Learning.
% The fix-selection abbreviations are: Human User Input (UI), Reliability Levels (RL), (MA), Minimality of Entailment Loss (MEL), (SML), (UML), (EK)
}
\label{tab:fixing}
\end{center}
\end{table}

% ABox repairing of a KG can also be seen as a special case of the data cleaning problem, where the TBox represents the schema and the constraints of a Knowledge Base and the ABox contains the data. 
% A variety of methods have been proposed for data repairing in the context of the established field of data cleaning, which consists of detecting and repairing errors in data. 
% An error in the data is a notion broader than a formal inconsistency, which is the topic of this work. However, in the context of qualitative data cleaning, where the errors violate specific constraints, rules, or patterns, inconsistencies can be handled as errors and vice versa.   
% A classification and overview of qualitative data cleaning approaches is provided by~\cite{Chu2016}.

\subsubsection{Generating fixes}
% A fix can be any set of actions that restores consistency in the KG. 
A fix restores consistency in a KG, i.e. leads to a repair, as defined in Definition~\ref{def:fix}. 
As regards ABox fixing, most works focus on the deletion of some ABox assertions~\cite{Bonatti2011,Melo2020,Baader2022}. KG fixing through both deletion and addition of assertions has also been investigated, for example in the context of satisfying constraints on minimum cardinality~\cite{PellissierTanon2019,Ahmetaj2022}. 
% https://www.w3.org/TR/shacl/#MinCountConstraintComponent
Adding assertions combined with deletions can also support the special case of ``update-based repairing''~\cite{Melo2020,Arnaout2022}. Contrary to ``deletion-based'' repairing, in update-based approaches, only parts of an assertion are altered in an effort to retain as much original information as possible. In the example of Figure~\ref{fig:KG_example} for instance, an update-based fix could include changing the subject of the \texttt{Bob} \texttt{belongsTo} \texttt{Robby} assertion into \texttt{Robby}, and the object into \texttt{Bob}, as presented in the fix of Example~\ref{fixExample}, retaining the \texttt{belongsTo} relation instead of removing it from the KG. In this regard, different alternative fixes may be available and the automated calculation of such alternative fixes is a first goal towards fixing. In the fix of Example~\ref{fixExample} for instance, any \texttt{Person} could be the subject of the assertion without raising inconsistency, not only \texttt{Bob}.   
Some important challenges in this task include the identification of alternative fixes that can indeed lead to consistency, as well as the efficiency of this process, as the number of alternatives to be considered as potential fixes can be large.

% \todo[inline]{Fan2019 here with certain fixes, but do not cover all the cases.}
Several methods in the data-cleaning literature rely on variations of the \algo{chase} algorithm~\shortcite{Benedikt2017} for reasoning which allows them to generate and apply update actions during the reasoning process, and develop a fix incrementally, without exhaustive consideration of all potential combinations of update actions~\shortcite{Geerts2020}.
% ~\cite{Arioua2018} (update-based~\cite{Wijsen2005})
% Cleaning data with Llunatic~\cite{Geerts2020}
Though such methods are usually not designed particularly for KGs, their application to KGs is often feasible, provided that an adequate transformation between the adopted formalizations is feasible. That is, the KG should be fully expressed in the language that each method supports. \citeauthory{Arioua2018}, for example, consider a TBox consisting of specific types of relational database constraints, namely Tuple-generating and conflict detection dependencies, which are less expressive than OWL.
In addition, graph-specific methods have also been proposed for the data-fixing task in the context of the graph-database literature. For example, $\mathit{Gfix}$ defines the notion of graph quality rules and introduces a graph-specific variation of the \algo{chase} algorithm which is used for deducing certain fixes on an inconsistent KG~\cite{Fan2019}.
% \item GQRs: Graph Quality Rules. ``Deducing certain fixes to graphs'' Considering de-duplication in parallel to inconsistency repairing~\cite{Fan2019}.
% \item ``A logic-based incremental approach to graph repair featuring delta preservation''~\cite{Schneider2021}

% \todo[inline]{Melo 2020 here for ML for fix generation, but do not guarantee consistency.}

\subsubsection{Choosing among fixes}
When alternative fixes are available, the choice of which of them to adopt is the other goal of KG fixing. 
Some works rely exclusively and directly on human user input for such choices~\cite{Arioua2018} but others attempt to automate this process. 
For instance, this can be done by employing the idea of active integrity constraints in the TBox, which is extended to include policies for consistency restoring, as overviewed by \citeauthory{FeuilladeHR20}.
Another way to automatically choose among fixes is by checking which parts of the ABox are more reliable, based on given reliability levels. \citeauthory{Fan2019} for example, consider a part of the ABox as ground truth and incrementally resolve potential inconsistencies with fixes that can be entailed by this part, consulting human users only in cases where such a fix is not available. 
In the example of Figure~\ref{fig:KG_example} for instance, even if the assertion \texttt{Robby} \texttt{belongsTo} \texttt{Bob} were part or could be entailed by a ground truth, still user input would be required for choosing among alternative fixes for updating \texttt{Robby} \texttt{hasName} \texttt{"2000"\string^\string^xsd:int}, such as replacing the object with \texttt{"2000"\string^\string^xsd:string} 
% \texttt{"2001"\string^\string^xsd:string}, 
or \texttt{"Robby"\string^\string^xsd:string}.
% , etc.
% \todo{shouldn't this be xsd:string?} Right, thanks!
Other works consider several reliability levels that prioritize different parts of the ABox, which may reflect, for example, the quality of their original resources.
% Bonatti2011
% \item ``How to select one preferred assertional-based repair from inconsistent and prioritized DL-lite knowledge bases?''~\cite{Benferhat2015} 
In this context, the fix that leads to a preferred repair can be chosen automatically, as the one that removes assertions of lower reliability level. 
% \item Automatic selection of the ``best'' repair to handle inconsistent triples. ``Certain'' vs ``best'' vs ``uncertain'' data cleaning based on whether and how the inconsistent triples are considered.~\cite{Benbernou2017}. The notion of ``best'' repair has also been extended into a temporal setting~\cite {Tahrat2021}. 
To make the computation of such repairs feasible, \citeauthory{Bonatti2011} consider the $O2R^-$ subset of rules in the Rule Language profile of OWL 2 (\textit{OWL 2 RL}\footnote{https://www.w3.org/TR/owl2-profiles/\#OWL\_2\_RL}), which is selected to ensure linear growth of assertional inferences with regard to the ABox size.
\citeauthory{Benferhat2015} focus on prioritized $\DLLite$ KBs~\shortcite{Calvanese2007}.   

Another way of choosing among alternative fixes is based on the minimality of update actions in a fix. \citeauthory{Ahmetaj2022}, for instance,
inspired by related work on database repairing, propose a method that identifies a minimal set of update actions on an RDF KG to conform to a set of constraints, expressed in the Shapes Constraint Language\footnote{https://www.w3.org/TR/shacl/} (SHACL). This conformation can be full, or in cases that this is not possible, it can be maximal, fixing as many assertions as possible.
% We presented an approach to repair a data graph so that it conforms to a set of SHACL constraints.
% \item ``Repairing SHACL Constraint Violations Using Answer Set Programming''~\cite{Ahmetaj2022}
% \item SHACTOR: Shapers extrACTOR from very large KGs. ``Improving the Quality of Large-Scale Knowledge Graphs with Validating Shapes''~\cite{Rabbani2023}
In a similar point of view, \citeauthory{Baader2022} focus on $\mathcal{EL}$\footnote{https://www.w3.org/2007/OWL/wiki/EL} KBs with inconsistency that involves some entailed information and rely on the minimality of entailment loss. That is, they choose the fixes that lead to the so-called ``optimal repairs'', where no conflicting information is entailed anymore, but the non-conflicting entailments are retained as much as possible.
% How to repair the ontology such that some ``unwanted consequences'' can no longer be deduced? 
%  \begin{itemize}
%     \item ``Query-Driven Repairing of Inconsistent DL-Lite Knowledge Bases''~\cite{Bienvenu2016} ``The aim is to find a set of ABox modifications (deletions and additions), called a repair plan, that addresses as many of the defects as possible''
%     \item Weakening assertions that generate inconsistent implications instead of removing them (``Making repairs in description logics more gentle''~\cite{Baader2018})
%     \item ``An optimal repair w.r.t. an entailment relation is one where the least amount of other (non-unwanted) consequences is removed.'', ``quantified ABoxes (qABoxes) [16], where in addition to the usual named individuals we also have anonymous objects, which are represented as (existentially quantified) variables.'' ~\cite{Baader2021,Baader2022}
% \end{itemize}

% Beyond human user input and known reliability levels, 
Machine Learning (ML) models have also been applied for KG fixing, mostly in the context of KG completion and error detection. 
% the primary aim is to minimize erroneous assertions and increase the veracity of the KG. C
Consistency checking is employed by some methods, mainly considering automatically extracted constraints as a means towards KG completion and error detection. But leading to a consistent KG considering a pre-existing TBox is often not guaranteed.
% % \item ``Neural Knowledge Base Repairs'' learns the repair actions directly from the edit history: human edits are still needed as history.~\cite{PellissierTanon2019,PellissierTanon2021}
% Pellissier Tanon \textit{et al.}~\citeyear{PellissierTanon2019,PellissierTanon2021} 
\citeauthor{PellissierTanon2019}~\citeyear{PellissierTanon2019,PellissierTanon2021} employ supervised ML approaches to learn specific correction rules from the history of resolving previous errors in an RDF KG by human experts. 
However, most approaches focus on unsupervised ML and learn to identify erroneous parts of the ABox based on frequent patterns. 
\citeauthory{Paulheim2014}, for example, consider type and property distributions in RDF KGs to identify missing types of entities and wrong property assertions that link entities of unexpected types. 
\citeauthory{Cheng2020} propose mining rules that can capture and fix different types of errors considering graph patterns as well, by introducing Graph Repairing Rules and Flexible Graph Repairing Rules. 
% Graph Repairing Rules (GRRs) and Flexible GRRs ($\delta$-GRRs). 

\citeauthory{Melo2020} mine patterns, by developing a decision tree classifier, and then translate them into SHACL constraints. They propose the development of decision-tree models that rely on path and type features on a KG to score the relation assertions as potentially erroneous, \ie as candidates for removal or correction. The correction is performed by updating the type of some entity with a type estimated by another ML model, or by replacing the entity with a ``similar'' one. The similarity between entities is estimated based on the Wikipedia disambiguation links and name string similarities. The resulting relation assertions are ranked according to the confidence of the ML model for them not being erroneous.
% , and the confidence assigned to  the ML model.  
% development of classifiers that rely on path and type features to identify erroneous relation assertions for removal      
% \item Entity-confusion errors (ML-based): 
% \begin{itemize}
%     \item ``An approach to correction of erroneous links in knowledge graphs''~\cite{Melo2017}, and
%     \item an extension (CoCKG) for SHACL relation constraints in RDF graphs~\cite{Melo2020}.
% \end{itemize}
%       \begin{itemize}
%           \item ``Strict and Flexible Rule-Based Graph Repairing''~\cite{Cheng2020}, and 
%           \item ``Rule-Based Graph Repairing: Semantic and Efficient Repairing Methods''~\cite{Cheng2018}
%       \end{itemize}
In the same direction, other methods rely on neural networks to estimate the correctness of specific assertions for fixing KGs, particularly neural embedding representations. \citeauthory{Abedini2022}, for instance, propose a method that employs KG embedding approaches to learn a model for estimating whether an assertion is true or not. Based on these estimations, they rank the alternative assertions for unique RDF relations, where an entity is related at most to one other entity. More recently, \citeauthory{Ye2023} combine both KG embedding representations and first-order logical rules with confidence weights for the same task.
%          \item EPCI: An Embedding Method for Post-Correction of Inconsistency in the RDF Knowledge Bases~\cite{Abedini2022} and Correction Tower~\cite{Abedini2020} 
% The different ML directions are not mutually exclusive, and some works combine some of them. 
%          \item ``Grier iteratively learns the graph embeddings with guidance from logical rules''~\cite{Ye2023}

Finally, beyond direct human-user input and patterns extracted from the KG itself, external knowledge sources (EK) have also been considered. 
% Aprosio \textit{et al.}~\citeyear{Aprosio2013}, for instance, use Wikipedia as a source for distant supervision for completing missing information.
\citeauthory{Melo2020}, for instance, exploit the Wikipedia disambiguation pages to estimate entity similarity, as already discussed above. 
\citeauthory{Gad-Elrab2019} exploit textual web resources to
collect supporting evidence for specific assertions and provide it to the human KG curators, based on rules expressed as Horn clauses, and \citeauthory{Arnaout2022} propose the use of probes on pretrained language models for choosing the right entity to replace
the object of an erroneous relation assertion.
%          \item Utilizing Language Model Probes for Knowledge Graph Repair~\cite{Arnaout2022}
% The aforementioned methods for fixing a KG with a reliable TBox are summarized in Table~\ref{tab:fixing}.
% , annotated with respective constraints considered and the approach adopted for fix selection. 
% \todo[inline]{overall comment on how is related to previous and next sections \& expressiveness in relation to detection.}

Analytical approaches usually rely on specific logic formalizations and optimization strategies regarding the fixes generated.
However, for choosing among alternative fixes they often require direct human user input or pre-defined priority/reliability levels which may not always be available in practice. ML-based approaches, on the other hand, usually rely on unsupervised techniques that exploit the KG itself without the need for special input. However, such approaches typically focus on the task of minimizing erroneous assertions, not necessarily leading to a consistent KG. The combination of the above directions is therefore very promising for the efficient fixing of inconsistent KGs. Still, for cases where choosing among the alternative fixes is not feasible or desired, there are approaches for reasoning while tolerating inconsistency, as discussed in the section that follows.   

% \item Assessing and Refining Mappings to RDF triples with a focus on the consistency of rules for the development of a KG. (``Our methodology ... (ii) automatically suggests mapping refinements based on the results of these quality assessments."~\cite{Dimou2015})
% \item QAPs: Quality Assessment Pattern representing quality metrics. ~\cite{Sejdiu2019} for common data quality problems.

% \newpage

\section{Inconsistency-Tolerant Reasoning}\label{sec:Paraconsistent}
% \todo[inline]{Lead:Angelos}

Inconsistent-tolerant reasoning emerges as a non-standard reasoning 
that refrains from fixing contradictions within the knowledge graph.
% This is useful when a permanent repair of a \gls{kg} is not desirable, 
% either because a copy of a \gls{kg} cannot be obtained or 
% because altering the original assertions is not allowed. 
Several approaches had defined different semantics that can tolerate
inconsistent \glspl{kg}. The approaches are summarized in
Table~\ref{tab:tolerating}.
While our focus is in reasoning on \glspl{kg}, the concept of 
inconsistency-tolerant reasoning is broader and has been applied in various other formalisms.
% In particular, there have been several related works 
% which explore inconsistent-tolerant semantics for \glspl{kb} 
% expressed as existential rules, another prominent class of ontology language
% (see, for example,
% \cite{lukasiewicz_inconsistency_2012}).

\begin{table}
\centering
\begin{tabular}{llll}
\toprule
Approach                                                  & Language &   Semantics  \\
\midrule
\shortciteA{lembo_inconsistency-tolerant_2010}      & $\DLLite_{\mathcal{A},\mathit{id},\mathit{den}}$    & IAR              \\
\shortciteA{rosati_evaluation_2012}                 & $\DLLite_\mathcal{A}$                               & IAR               \\
\shortciteA{bienvenu_2014} \checkmark               & $\DLLite_\mathcal{R}$                               & IAR, brave, AR    \\
\shortciteA{orbits} \checkmark                      & $\DLLite_\mathcal{R}$                               & preferred repairs \\
\shortciteA{du_weight-based_2013}                   & $\mathcal{SHIQ}$                                    & Weight-based AR   \\
\shortciteA{tsalapati_efficient_2016} \checkmark    & $\DLLite_{\mathcal{R},\sqcap}$                      & ICAR, IAR \\
\shortciteA{trivela_framework_2018}                 & $\mathcal{EL}_{\bot n r}$                           & IAR       \\
\shortciteA{trivela_query_2019}                     & $\mathcal{EL}_{\bot n r}$                           & ICAR       \\
\shortciteA{zhang_inconsistency-tolerant_2014}      & QC-$\mathcal{SHOIN}(\mathrm{D})$                    & Quasi-classical \\
\shortciteA{jung_ontology-based_2012}               & $\mathcal{ELI}$                                     & Probabilistic  \\
\shortciteA{benferhat_non_2015}                     & $\DLLite_{\mathcal{R}}$                             & Possibilistic \\
\bottomrule
\end{tabular}
\caption{Overview of inconsistent-tolerant approaches.  \checkmark indicates a publicly available implementation.}
\label{tab:tolerating}
\end{table}

\subsection{Repair-based Reasoning}

% One of the main ideas 
A prominent concept involves performing reasoning as if the knowledge graph is
repaired. In other words, the reasoning is happening in a repair of the knowledge 
graph. The notion of repair-based reasoning has been initially
introduced by \citeauthory{arenas_consistent_1999} for query answering 
over relational databases (actually, they called their approach consistent query answering). 
This idea was later adapted by \citeauthory{lembo_inconsistency-tolerant_2010} for ontological reasoning,
leading to the development of the so-called \gls{AR} semantics
for various \glspl{DL} where AR stands for ABox repair.
Most of the repair-based reasoning approaches operate under the assumption 
that the TBox is reliable and the only possible kind of repair is an ABox repair.
However, this is not the only solution and it has also been generalized 
to handle the case where both TBox axioms and ABox assertions 
may be erroneous~\cite{eiter_generalized_2016}.

The primary challenges for repair-based reasoning involve selecting an
appropriate repair of the \gls{kg} and determining how it can be computed
dynamically. Most related approaches differentiate based on these criteria.

\subsubsection{Selecting Repairs}

In the \gls{AR} semantics, it is assumed that all repairs 
are equally appropriate and a conclusion is entailed only if it holds true across \emph{all possible repairs}.
The intuition is that, in the absence of further information, we cannot identify the ``correct''
repair, so we only consider an answer to be entailed if it can be obtained from
every ABox repair. Conversely, the \emph{brave} semantics suggests that 
a conclusion is entailed if it is obtained by any of the possible ABox repairs.

Besides the AR semantics, other types of repairs have been proposed.
For example, the Closed ABox Repairs (CAR) semantics has been proposed in order to overcome some counterintuitive 
cases of the AR semantics. In the AR semantics it matters whether an assertion is explicitly 
defined in the ABox or inferred using the TBox, a distinction that can lead to  
examples where the conclusions of two very similar \glspl{kg} differ unexpectedly. 
The CAR semantics addresses this issue; the ABox is enriched by enclosing all the consistent conclusions. 
This requires a modified closure operator to be applied to the original inconsistent ABox
that produces only the conclusions that are based on consistent information.
The modified ABox is usually called closed ABox and the resulting ABox repairs are called CAR.
A great overview of the different repair-based semantics can be found 
in~\cite{bienvenu_short_2020}.

%
%\paragraph{AR semantics and friends}

%
%\paragraph{Need to approximate AR semantics}
Determining the set of consistent answers under the AR or CAR semantics is
known to be a hard problem, even for very simple DLs.
For example, deciding whether an assertion is entailed by a $\DLLite_\mathit{core}$
KG under the AR semantics is \coNP-complete with respect to data complexity~\cite{rosati_complexity_2011}. 
For this reason, several
other semantics that approximate AR or CAR semantics have been proposed 
aiming at restoring tractability. In IAR (Intersection of AR) semantics~\cite{lembo_inconsistency-tolerant_2015}, 
an answer is considered to be entailed only if it can be obtained 
by the intersection of all ABox repairs and in a completely analogous way, 
the (Intersection of CAR) ICAR semantics~\cite{lembo_inconsistency-tolerant_2015} 
use the intersection of all closed ABox repairs. 
In some cases, computing the intersection of ABox repairs is easier than 
computing all ABox repairs. Roughly speaking, the computation of a repair involves 
a non-deterministic choice of the assertion in a conflict set to be removed while in the
computation of the intersection of repair all assertions in a conflict set should be removed.
In fact, reasoning is tractable for 
$\DLLite_\mathcal{A}$ under the IAR or ICAR semantics but remains intractable 
for more expressive DLs~\cite{rosati_complexity_2011}.

%\paragraph{Selection of repairs based on preference}
Some works consider other qualitative information for selecting the appropriate repairs,
for example, the relative reliability of ABox assertions. If there is information
regarding the reliability of assertions, one can utilize it to identify
\emph{preferred repairs} and enhance the accuracy of query results. A preference 
relation between repairs can be expressed in several ways. One criterion is to 
define preferred repairs as the cardinality-maximal ones~\cite{bienvenu_2014}.
A more elaborate criterion is to assign weights to assertions, for example based on the 
trustworthiness of the data source. In that case the preferred repairs can be defined 
as those with the highest overall weight~\cite{du_weight-based_2013}. A different approach 
is to avoid the assignment of weights altogether and define a partial preference ordering 
among assertions. This ordering induces a natural partitioning of the ABox where assertions
of the same preference belong to the same ABox subset~\cite{elect}. 
In~\cite{orbits} 
several preferential relations over the ABox have been studied and a general framework 
has been proposed for computing the preferred repairs. Unsurprisingly, reasoning over 
the preferred repairs is still intractable. Preferences can also be defined on rules.
For example, \citeauthory{calautti_preference-based_2022} proposed preferential rules 
in the setting of existential rules for defining user preferences among repairs.

\subsubsection{Computing Repairs}

There are several systems that implement reasoning under the IAR or ICAR
semantics mainly for less expressive DLs such as $\DLLite$. One of the first
implementations is QuID~\cite{rosati_evaluation_2012} that performs conjunctive
query answering under the IAR semantics in an extension of
$\DLLite_\mathcal{A}$. QuID has implemented and compared three different
approaches: ABox annotation, ABox cleaning and first-order query rewriting. The
first two techniques require a preprocessing of the entire ABox. In ABox
annotation each assertion is marked as either safe or problematic where
problematic assertions are exactly those that belong to a conflict set. Given a
query, the ABox is modified to use only safe assertions. On the other hand, ABox
cleaning removes all problematic assertions in order to compute the intersection of
repairs. Both of these techniques involve data modification, which contradicts
the primary motivation of inconsistent-tolerant reasoning. Lastly, 
first-order query rewriting is a technique that rewrites the initial query
to filter out any answer that depends on assertions not included in all repairs.
This is done by adding extra first-order conditions to the initial query by 
consulting the TBox axioms.
The main advantage of this technique is that the resulting query can be
executed directly to the inconsistent \gls{kg}.
As a result, the intersection of repairs will never be explicitly computed but only
the part that is necessary for answering the specific query. Moreover, in
principle, the resulting query is a first-order formula and can be expressed as an SQL query 
and be executed from efficient relational database systems. 
In practice, however,
transformed queries tend to be large and inefficient. 

A different technique that is 
using a Datalog rewriting instead of a first-order rewriting is proposed by
\citeauthory{tsalapati_efficient_2016} for
implementing the ICAR semantics of $\DLLite_\mathcal{R}$. Instead of incorporating the TBox to the
query, the TBox axioms are translated into Datalog rules. The 
translation to Datalog is guaranteed if the TBox axioms are in $\DLLite_\mathcal{R}$.
Then, a Datalog engine can be used to materialize the consequences more effectively. 
The combination of query rewriting and materialization,
together with some other optimizations, achieve comparatively faster times than other
systems. Other works that target more expressive DLs also use Datalog
rewritings. For example, \citeauthory{trivela_framework_2018,trivela_query_2019} target
$\mathcal{EL}_{\bot nr}$. Even though they target a more expressive \gls{DL},
their algorithm does not always terminate since termination depends on the form of the TBox axioms.

% \paragraph{Approximations}
In addition to techniques for implementing the IAR and ICAR semantics, there are 
related works for implementing the AR semantics. Since reasoning with AR semantics is intractable,
\citeauthory{bienvenu_rosati_2013} proposed an approach for
approximating the query answers using an upper and a lower approximation. They
defined a set of different semantics that approximate AR semantics 
but are easier to compute. Iteratively, one can compute tighter
approximations until the lower and upper approximations coincide.
\citeauthory{bienvenu_2014} use as lower approximation the IAR semantics and
upper approximation the brave semantics, leaving, in general,
a ``gap'' (\ie the difference between the two approximations) of answers.
The rest of the AR answers are identified by
verifying each tuple in the gap by a SAT solver via constructing an instance of
UNSAT. Experiments show that despite its intractable data complexity, it is
feasible to compute query answers under the AR semantics, thanks in part to the
fact that many AR answers can be identified using the tractable IAR semantics.

% \subsection{Defeasible Reasoning}

% Defeasible reasoning can also address inconsistencies. A key characteristic of
% these logics is the inclusion of ``defeasible'' axioms within the language. This
% is typically represented by a new defeasible implication operator.

\subsection{Many-valued Reasoning}

An alternative way to get meaningful answers from an inconsistent theory is to
adopt paraconsistent logics which have more than two truth values. Roughly
speaking, contradictory formulas are labeled as over-defined, \ie a third truth
value that signifies that we can prove both their truth and falsity, resulting
in classical inconsistency. Such paraconsistent logics enable meaningful reasoning with
variables that are not directly involved in a contradiction. Several
paraconsistent logics have been developed for \glspl{DL} (see,
\cite{kamide_comparison_2013} for a comparison).
In those semantics, negation is not interpreted
as just the set complement but as an operator on the many truth values. Some
approaches allow multiple versions of the implication and negation operations.
For example, \citeauthory{zhang_inconsistency-tolerant_2014} proposed an extension 
of a \gls{DL}, so called QCDL (Quasi-classical DL), by introducing a new negation operator, called 
QC negation. They implemented the semantics of QCDL by reducing QCDL to 
classical \gls{DL} and then reason using a classical \gls{DL} reasoner.

% The original monotonic form of these logics tends to be overly cautious, yielding
% fewer logical consequences in the case of a consistent \gls{kg} compared
% to classical reasoning. To address this, a preference for minimally-inconsistent
% interpretations is employed, prioritizing interpretations with the least
% over-defined variables.

% \subsubsection{Inconsistencies as uncertainty}

Inconsistency is also closely related to uncertainty. This association stems
from the fact that uncertainty can be a result of inconsistency. For example,
responses to a query, when viewed through various inconsistency-tolerant
frameworks, may vary in perceived certainty. Additionally, inconsistency can be
mitigated to a certain degree by acknowledging uncertainty in certain pieces of
knowledge. This allows for the coexistence of two contradictory
yet uncertain statements.

There has been a number of proposals for probabilistic DLs. Probabilistic
KBs~\cite{jaeger_probabilistic_1994} associate probabilities to the ABox
assertions, and the TBox contains both classical and probabilistic
axioms that take the form of conditional probability which
express statistical probabilities. 
\citeauthory{jung_ontology-based_2012} introduced probabilistic ABoxes that
instead of associating ABox assertions with probabilities it associates them 
to probabilistic events. Roughly speaking, a set of probabilistic events are expressions
with attached probabilities that can be either atoms or can be composed by simpler probabilistic events.
In case of compound events, the attached probability is inferred.

In possibilistic \glspl{DL}~\cite{qi_possibilistic_2007}, each assertion 
is associated with a degree of certainty, and the $\alpha$-cut of a knowledge graph 
consists of the assertions of degree greater than $\alpha$.
The inconsistency degree $\gamma$ of a knowledge graph is the maximum degree 
such that the $\gamma$-cut of the \gls{kg} is inconsistent. In other words, 
for strictly greater degrees than $\gamma$, the corresponding cut is consistent.
Then, we consider the plausible consequence of the \gls{kg}, \ie the consequences 
of the $\delta$-cut where $\delta > \gamma$.
\citeauthory{benferhat_non_2015} investigated inconsistency-tolerant
semantics in possibilistic \DLLite, which extend the plausible consequences
by adding to the $\delta$-cut the assertions that are free of
conflicts, or conflicted only by assertions of lower degree of certainty.

\section{Summary and Open Challenges}\label{sec:Conclusions}
% \todo[inline]{Lead:Tasos}
% \todo[inline]{Brief summary}
% \todo[inline]{Open challenges}
We focused on the timely topic of reasoning on inconsistent KGs. 
First, we introduced some basic notions and terms about inconsistent KGs, providing definitions and examples.
Then, we classified the relevant literature into the following tasks: 
\begin{enumerate*}[label=\alph*)]
\item the detection of inconsistency in KGs,
\item the fixing of inconsistent KGs, and 
\item reasoning in KGs in the presence of inconsistency. 
\end{enumerate*}

A variety of methods has been proposed for inconsistency detection in KGs, which is coming up with inconsistency explanations. These methods often employ classical DL reasoners, but time complexity might be prohibiting in cases of large KGs.  Scalability can be achieved either by splitting the reasoning process into smaller tasks that can be performed in parallel, by reducing the expressivity of the supported languages, or by admitting approximate methods as alternatives. 
Such methods of approximate reasoning often rely on machine-learning models that provide predictions that are not guaranteed to be correct, sacrificing completeness for scalability.  
% An overview of relevant works organized based on these dimensions is provided in Table~\ref{tab:overview}.

Fixing an inconsistent KGs involves altering its elements. Our focus here was on altering the Abox, as it is considered the largest and most error-prone part of several real-world KGs.  
Scalability into large KGs is a challenge for KG fixing as well and optimizations in the reasoning process, parallelization, and Machine-learning approaches have also been proposed for this task. 
An additional challenge in KG fixing is choosing among alternative KG fixes. In this direction, the methods often rely on different sources ranging from direct user input and external knowledge sources to predefined reliability levels and estimations produced by machine-leaning models.  

When choosing among the alternative fixes is not feasible, several approaches define different semantics that can tolerate inconsistent reasoning.
A prominent concept is repair-based reasoning, which involves performing reasoning as if the knowledge graph is repaired. A major challenge in this direction is selecting an appropriate repair of the KG and determining how it can be computed dynamically. This can be done considering \emph{all possible repairs}, as done in the \gls{AR} semantics or enriching the ABox with all consistent conclusions, as done in the CAR semantics. However, considering the intersection of \gls{AR} and CAR, under the semantics IAR and ICAR, can be computationally favorable.  
Finally, meaningful answers from an inconsistent KG para-consistent can also be derived by adopting many-valued logics that have more than two truth values or associating inconsistency with uncertainty through probabilistic reasoning. 

\gls{AR} semantics, it is assumed that all repairs 
% are equally appropriate and a conclusion is entailed only if it holds true across \emph{all possible repairs}

% Beyond presenting how and in which cases different fields and specific methods are related to reasoning on inconsistent KGs, we also highlight open challenges and future directions in addressing this problem and any particular subtasks of it.
From our study, a number of open challenges has come up. 
Regarding inconsistency detection:
\begin{itemize}
    \item 
    Readily available methods for obtaining complete sets of inconsistency explanations do not scale for large inputs, as is the case of most modern KGs. Thus, there is a need for tractable inconsistency detection open-source implementations. Most probably, these will rely on KG splitting or modularization techniques, so that parallel computational infrastructure can allow horizontal scalability.
    \item To achieve scalability, researchers tend to reduce the expressivity of supported languages, restraining this way the applicability of the methods for more complex real-world KGs. In face of this, the support for more expressive DL languages is an important requirement. One example is $\mathcal{SROIQ}(\mathbf{D})$, the DL equivalent to OWL 2 DL, which is the semantic representation language used widely in the modern web.
    \item Apart from reducing the supported language expressivity, scalability can be achieved by the incorporation of approximate techniques. However, the robustness of the approximation is not always clearly defined. To this end, the development of even faster alternatives for application in cases where the completeness of results is not necessary, should be also accompanied with proper bounds and guarantees, e.g. regarding soundness and completeness.
\end{itemize}
%Although module-splitting approaches such as that of~\cite{tran2020fast} can
%perform inconsistency detection on KGs efficiently and exploit the power of
%parallelization, there is still room for improvement in the expressiveness of
%the supported logic.

As regards the task of inconsistent KG fixing, there are several open challenges and promising directions for future research:
\begin{itemize}
    \item The implementation availability of KG-fixing methods is limited, with less than half of the studies presented in Table~\ref{tab:fixing} being accompanied by a publicly available implementation. 
    % direction
    Providing the implemented methods as open-source repositories would facilitate the reproducibility of results, the comparison of the methods, and their extension, allowing faster advancement in the field.     
    \item The constraints considered in the literature related to KG fixing range from Horn clauses and relational database constraints to SHACL and more specialized rule types, such as Graph-quality rules, RDF-correction rules, and Graph-repairing rules.
    % , such as the ones of~\citeauthory{Bogaerts2022,}. 
    % direction
    This variability highlights the lack of a unified formalization covering all the KG-fixing-related needs and the necessity of studies and tools for comparing and translating constraint sets into different formalizations.  
    \item KG-fixing methods often rely on different input resources for fix selection, ranging from direct user input and pre-defined reliability levels to patterns extracted from the KG itself, external KBs, and even pre-trained Language Models. Table~\ref{tab:fixing} reveals that only a few of the discussed methods combine two fix-selection resources and none of them more than two.
    % direction
    Hence, exploiting different fix-selection resources together seems an unexplored and interesting direction that could benefit from the potential complementarity of resources.    
    % , semantic web standards, such as OWL 2, $\mathcal{EL}$, and SHACL 
    \item Still, an important challenge for analytical KG-fixing methods remains their efficiency on web-scale KGs, where the calculation and selection of fixes often require the consideration of alternative update actions and performing consistency checks on the respective KGs. 
    ML-based methods, on the other hand, often overcome the computational obstacle of analytical calculation and testing of fixes by focusing on empirical predictions, at the expense of not guaranteeing consistency.
    % For this reason, most methods limit the expressivity of the supported constraints and/or develop the fixes incrementally, guided by user input, reliability levels, etc.
    % direction
    Therefore, the combination of analytical and ML-based approaches is a promising direction that can lead to KG consistency efficiently, guided by veracity estimations of error detection methods. 
\end{itemize}

Lastly, there are also several open challenges for inconsistency-tolerant reasoning:
\begin{itemize}
      % \paragraph{Expressivity}
\item It is evident that the expressivity of the language is a significant
      barrier to devising an efficient implementation. Most of the work
      presented in Section~\ref{sec:Paraconsistent} study less expressive DL
      fragments for this reason. Implementing efficiently inconsistent-tolerant 
      semantics for more expressive DLs is still an open challenge. 
      In this context, novel approximate techniques leveraging
      scalable systems, such as Datalog systems, may be particularly valuable.
      Furthermore, advancements in SAT (Satisfiability Testing) and ASP (Answer
      Set Programming) solvers for addressing computationally challenging
      problems could be essential in achieving this objective.

      % \paragraph{Scalability}
\item Reasoning with web-scale Knowledge Graphs is crucial for applications
      requiring the extraction of insights and the answering of complex queries from
      massive datasets. As the size of these datasets increases to web scale, even
      reasoning within consistent knowledge graphs becomes challenging. One approach
      is to consider \DLLite{} languages, which offer a good balance between
      expressivity and computational complexity. Although inconsistent-tolerant
      semantics exist for several \DLLite{} languages, implementing a web-scale
      reasoner remains an open challenge.

% \item There are several semantics that can potentially be used for
%       inconsistent-tolerant reasoning over \glspl{kg}. 
%       However, the advantages and disadvantages of each remain unclear.
\end{itemize}

% \newpage
% comment out acknowledgments for submission
% \iffalse
\iftrue
\section*{Acknowledgments}
This work has been supported by the ENEXA project, funded by the European Union's Horizon 2020 research and innovation programme, under grant agreement No 101070305.
\fi

\bibliography{survey-full,cqa,probdl}

\begin{thebibliography}{}

\bibitem[\protect\BCAY{Abedini, Keyvanpour,\ \BBA\ Menhaj}{Abedini
  et~al.}{2022}]{Abedini2022}
Abedini, F., Keyvanpour, M.~R., \BBA\ Menhaj, M.~B. \BBOP2022\BBCP.
\newblock \BBOQ {EPCI: An Embedding Method for Post-Correction of Inconsistency
  in the RDF Knowledge Bases}\BBCQ\
\newblock {\Bem IETE Journal of Research}, {\Bem 68\/}(2), 1043--1055.

\bibitem[\protect\BCAY{Abu-Salih}{Abu-Salih}{2021}]{Abu-Salih2021}
Abu-Salih, B. \BBOP2021\BBCP.
\newblock \BBOQ {Domain-specific knowledge graphs: A survey}\BBCQ\
\newblock {\Bem J. Comput. Netw. Commun.}, {\Bem 185}.

\bibitem[\protect\BCAY{Ahmetaj, David, Polleres,\ \BBA\ {\v{S}}imkus}{Ahmetaj
  et~al.}{2022}]{Ahmetaj2022}
Ahmetaj, S., David, R., Polleres, A., \BBA\ {\v{S}}imkus, M. \BBOP2022\BBCP.
\newblock \BBOQ {Repairing SHACL Constraint Violations Using Answer Set
  Programming}\BBCQ\
\newblock In {\Bem ISWC 2022}, \lowercase{\BVOL}\ 13489 LNCS, \BPGS\ 375--391.

\bibitem[\protect\BCAY{Arenas, Bertossi,\ \BBA\ Chomicki}{Arenas
  et~al.}{1999}]{arenas_consistent_1999}
Arenas, M., Bertossi, L., \BBA\ Chomicki, J. \BBOP1999\BBCP.
\newblock \BBOQ Consistent query answers in inconsistent databases\BBCQ.
\newblock {PODS} '99.

\bibitem[\protect\BCAY{Arioua\ \BBA\ Bonifati}{Arioua\ \BBA\
  Bonifati}{2018}]{Arioua2018}
Arioua, A.\BBACOMMA\  \BBA\ Bonifati, A. \BBOP2018\BBCP.
\newblock \BBOQ {User-guided Repairing of Inconsistent Knowledge Bases}\BBCQ\
\newblock {\Bem EDBT}, {\Bem 2018-March}, 133--144.

\bibitem[\protect\BCAY{Arnaout, Stepanova, Razniewski,\ \BBA\ Weikum}{Arnaout
  et~al.}{2022}]{Arnaout2022}
Arnaout, H., Stepanova, D., Razniewski, S., \BBA\ Weikum, G. \BBOP2022\BBCP.
\newblock \BBOQ {Utilizing Language Model Probes for Knowledge Graph
  Repair}\BBCQ\
\newblock In {\Bem WWW '22 Companion}, \lowercase{\BVOL}~1. Association for
  Computing Machinery.

\bibitem[\protect\BCAY{Artale, Calvanese, Kontchakov,\ \BBA\
  Zakharyaschev}{Artale et~al.}{2009}]{dllite}
Artale, A., Calvanese, D., Kontchakov, R., \BBA\ Zakharyaschev, M.
  \BBOP2009\BBCP.
\newblock \BBOQ The {DL-Lite} family and relations\BBCQ\
\newblock {\Bem J. Artif. Int. Res.}, {\Bem 36\/}(1), 1–69.

\bibitem[\protect\BCAY{Auer, Bizer, Kobilarov, Lehmann, Cyganiak,\ \BBA\
  Ives}{Auer et~al.}{2007}]{Auer2007}
Auer, S., Bizer, C., Kobilarov, G., Lehmann, J., Cyganiak, R., \BBA\ Ives, Z.
  \BBOP2007\BBCP.
\newblock \BBOQ {DBpedia: A Nucleus for a Web of Open Data}\BBCQ\
\newblock In {\Bem The Semantic Web}, \lowercase{\BVOL}\ 4825, \BPGS\ 722--735.

\bibitem[\protect\BCAY{Baader, Calvanese, McGuiness,\ \BBA\ Nardi}{Baader
  et~al.}{2007}]{baader2003description}
Baader, F., Calvanese, D., McGuiness, D.~L., \BBA\ Nardi, D.\BEDS.
  \BBOP2007\BBCP.
\newblock {\Bem The Description Logic Handbook: Theory, Implementation and
  Applications}.
\newblock Cambridge University Press.

\bibitem[\protect\BCAY{Baader, Koopmann, Kriegel,\ \BBA\ Nuradiansyah}{Baader
  et~al.}{2022}]{Baader2022}
Baader, F., Koopmann, P., Kriegel, F., \BBA\ Nuradiansyah, A. \BBOP2022\BBCP.
\newblock \BBOQ {Optimal ABox Repair w.r.t. Static EL TBoxes: From Quantified
  ABoxes Back to ABoxes}\BBCQ\
\newblock In {\Bem ESWC 2022}, \lowercase{\BVOL}\ 3263, \BPGS\ 130--146.

\bibitem[\protect\BCAY{Belabbes, Benferhat,\ \BBA\ Chomicki}{Belabbes
  et~al.}{2019}]{elect}
Belabbes, S., Benferhat, S., \BBA\ Chomicki, J. \BBOP2019\BBCP.
\newblock {\Bem Elect: An Inconsistency Handling Approach for Partially
  Preordered Lightweight Ontologies}, \BPG\ 210–223.
\newblock Springer International Publishing.

\bibitem[\protect\BCAY{Benedikt, Konstantinidis, Mecca, Motik, Papotti,
  Santoro,\ \BBA\ Tsamoura}{Benedikt et~al.}{2017}]{Benedikt2017}
Benedikt, M., Konstantinidis, G., Mecca, G., Motik, B., Papotti, P., Santoro,
  D., \BBA\ Tsamoura, E. \BBOP2017\BBCP.
\newblock \BBOQ {Benchmarking the Chase}\BBCQ\
\newblock In {\Bem SIGMOD/PODS'17}, \lowercase{\BVOL}\ Part F1277, \BPGS\
  37--52, New York, NY, USA. ACM.

\bibitem[\protect\BCAY{Benferhat, Bouraoui,\ \BBA\ Tabia}{Benferhat
  et~al.}{2015a}]{Benferhat2015}
Benferhat, S., Bouraoui, Z., \BBA\ Tabia, K. \BBOP2015a\BBCP.
\newblock \BBOQ {How to select one preferred assertional-based repair from
  inconsistent and prioritized DL-lite knowledge bases?}\BBCQ\
\newblock In {\Bem IJCAI}, \lowercase{\BVOL}\ 2015-Janua, \BPGS\ 1450--1456.

\bibitem[\protect\BCAY{Benferhat, Bouraoui,\ \BBA\ Tabia}{Benferhat
  et~al.}{2015b}]{benferhat_non_2015}
Benferhat, S., Bouraoui, Z., \BBA\ Tabia, K. \BBOP2015b\BBCP.
\newblock \BBOQ Non {Defeated}-{Based} {Repair} in {Possibilistic} {DL}-{Lite}
  {Knowledge} {Bases}\BBCQ\
\newblock In {\Bem {FLAIRS} 2015}. AAAI Press.

\bibitem[\protect\BCAY{Bienvenu}{Bienvenu}{2020}]{bienvenu_short_2020}
Bienvenu, M. \BBOP2020\BBCP.
\newblock \BBOQ A {Short} {Survey} on {Inconsistency} {Handling} in
  {Ontology}-{Mediated} {Query} {Answering}\BBCQ\
\newblock {\Bem KI - Künstliche Intelligenz}, {\Bem 34\/}(4).

\bibitem[\protect\BCAY{Bienvenu\ \BBA\ Bourgaux}{Bienvenu\ \BBA\
  Bourgaux}{2022}]{orbits}
Bienvenu, M.\BBACOMMA\  \BBA\ Bourgaux, C. \BBOP2022\BBCP.
\newblock \BBOQ Querying inconsistent prioritized data with {ORBITS:}
  algorithms, implementation, and experiments\BBCQ\
\newblock In Kern{-}Isberner, G., Lakemeyer, G., \BBA\ Meyer, T.\BEDS, {\Bem
  Proceedings of the 19th International Conference on Principles of Knowledge
  Representation and Reasoning, {KR} 2022, Haifa, Israel, July 31 - August 5,
  2022}.

\bibitem[\protect\BCAY{Bienvenu, Bourgaux,\ \BBA\ Goasdou{\'{e}}}{Bienvenu
  et~al.}{2014}]{bienvenu_2014}
Bienvenu, M., Bourgaux, C., \BBA\ Goasdou{\'{e}}, F. \BBOP2014\BBCP.
\newblock \BBOQ Querying inconsistent description logic knowledge bases under
  preferred repair semantics\BBCQ\
\newblock In {\Bem AAAI'14}, \BPGS\ 996--1002. {AAAI} Press.

\bibitem[\protect\BCAY{Bienvenu\ \BBA\ Rosati}{Bienvenu\ \BBA\
  Rosati}{2013}]{bienvenu_rosati_2013}
Bienvenu, M.\BBACOMMA\  \BBA\ Rosati, R. \BBOP2013\BBCP.
\newblock \BBOQ Tractable approximations of consistent query answering for
  robust ontology-based data access\BBCQ\
\newblock In {\Bem IJCAI'13}, IJCAI '13. AAAI Press.

\bibitem[\protect\BCAY{Bonatti, Hogan, Polleres,\ \BBA\ Sauro}{Bonatti
  et~al.}{2011}]{Bonatti2011}
Bonatti, P.~A., Hogan, A., Polleres, A., \BBA\ Sauro, L. \BBOP2011\BBCP.
\newblock \BBOQ {Robust and scalable Linked Data reasoning incorporating
  provenance and trust annotations}\BBCQ\
\newblock {\Bem J. Web Semant.}, {\Bem 9\/}(2), 165--201.

\bibitem[\protect\BCAY{Bordes, Usunier, Garcia-Duran, Weston,\ \BBA\
  Yakhnenko}{Bordes et~al.}{2013}]{NIPS2013_1cecc7a7}
Bordes, A., Usunier, N., Garcia-Duran, A., Weston, J., \BBA\ Yakhnenko, O.
  \BBOP2013\BBCP.
\newblock \BBOQ Translating embeddings for modeling multi-relational data\BBCQ\
\newblock In Burges, C., Bottou, L., Welling, M., Ghahramani, Z., \BBA\
  Weinberger, K.\BEDS, {\Bem Advances in Neural Information Processing
  Systems}, \lowercase{\BVOL}~26. Curran Associates, Inc.

\bibitem[\protect\BCAY{Calautti, Greco, Molinaro,\ \BBA\ Trubitsyna}{Calautti
  et~al.}{2022}]{calautti_preference-based_2022}
Calautti, M., Greco, S., Molinaro, C., \BBA\ Trubitsyna, I. \BBOP2022\BBCP.
\newblock \BBOQ Preference-based inconsistency-tolerant query answering under
  existential rules\BBCQ\
\newblock {\Bem Artif. Intell.}, {\Bem 312\/}(C).

\bibitem[\protect\BCAY{Calvanese, {De Giacomo}, Lembo, Lenzerini,\ \BBA\
  Rosati}{Calvanese et~al.}{2007}]{Calvanese2007}
Calvanese, D., {De Giacomo}, G., Lembo, D., Lenzerini, M., \BBA\ Rosati, R.
  \BBOP2007\BBCP.
\newblock \BBOQ {Tractable Reasoning and Efficient Query Answering in
  Description Logics: The DL-Lite Family}\BBCQ\
\newblock {\Bem Journal of Automated Reasoning}, {\Bem 39\/}(3), 385--429.

\bibitem[\protect\BCAY{Chen, Deng, Zhang, Xu, Li,\ \BBA\ Kharlamov}{Chen
  et~al.}{2021}]{chen2021neural}
Chen, H., Deng, S., Zhang, W., Xu, Z., Li, J., \BBA\ Kharlamov, E.
  \BBOP2021\BBCP.
\newblock \BBOQ Neural symbolic reasoning with knowledge graphs: Knowledge
  extraction, relational reasoning, and inconsistency checking\BBCQ\
\newblock {\Bem Fundamental Research}, {\Bem 1\/}(5), 565--573.

\bibitem[\protect\BCAY{Chen, Jia,\ \BBA\ Xiang}{Chen et~al.}{2020}]{Chen2020}
Chen, X., Jia, S., \BBA\ Xiang, Y. \BBOP2020\BBCP.
\newblock \BBOQ {A review: Knowledge reasoning over knowledge graph}\BBCQ\
\newblock {\Bem Expert Systems with Applications}, {\Bem 141}, 112948.

\bibitem[\protect\BCAY{Cheng, Chen, Yuan, Wang, Li,\ \BBA\ Jin}{Cheng
  et~al.}{2020}]{Cheng2020}
Cheng, Y., Chen, L., Yuan, Y., Wang, G., Li, B., \BBA\ Jin, F. \BBOP2020\BBCP.
\newblock \BBOQ {Strict and Flexible Rule-Based Graph Repairing}\BBCQ\
\newblock {\Bem IEEE Trans. Knowl. Data Eng.}, {\Bem 34\/}(7), 1--1.

\bibitem[\protect\BCAY{de~Groot, Raad,\ \BBA\ Schlobach}{de~Groot
  et~al.}{2021}]{de2021analysing}
de~Groot, T., Raad, J., \BBA\ Schlobach, S. \BBOP2021\BBCP.
\newblock \BBOQ Analysing large inconsistent knowledge graphs using
  anti-patterns\BBCQ\
\newblock In {\Bem The Semantic Web: 18th International Conference, ESWC 2021,
  Virtual Event, June 6--10, 2021, Proceedings 18}, \BPGS\ 40--56. Springer.

\bibitem[\protect\BCAY{Donini}{Donini}{2007}]{Donini_2007}
Donini, F.~M. \BBOP2007\BBCP.
\newblock {\Bem Complexity of Reasoning}, \BPG\ 105–148.
\newblock Cambridge University Press.

\bibitem[\protect\BCAY{Du, Qi,\ \BBA\ Shen}{Du
  et~al.}{2013}]{du_weight-based_2013}
Du, J., Qi, G., \BBA\ Shen, Y.-D. \BBOP2013\BBCP.
\newblock \BBOQ Weight-based consistent query answering over inconsistent
  $\mathcal{SHIQ}$ knowledge bases\BBCQ\
\newblock {\Bem Knowledge and Information Systems}, {\Bem 34\/}(2).

\bibitem[\protect\BCAY{Ehrlinger\ \BBA\ W{\"o}{\ss}}{Ehrlinger\ \BBA\
  W{\"o}{\ss}}{2016}]{ehrlinger2016towards}
Ehrlinger, L.\BBACOMMA\  \BBA\ W{\"o}{\ss}, W. \BBOP2016\BBCP.
\newblock \BBOQ Towards a definition of knowledge graphs.\BBCQ\
\newblock {\Bem SEMANTiCS (Posters, Demos, SuCCESS)}, {\Bem 48\/}(1-4), 2.

\bibitem[\protect\BCAY{Eiter, Lukasiewicz,\ \BBA\ Predoiu}{Eiter
  et~al.}{2016}]{eiter_generalized_2016}
Eiter, T., Lukasiewicz, T., \BBA\ Predoiu, L. \BBOP2016\BBCP.
\newblock \BBOQ Generalized consistent query answering under existential
  rules\BBCQ\
\newblock In {\Bem KR'16}, {KR}'16, Cape Town, South Africa. AAAI Press.

\bibitem[\protect\BCAY{Fan, Lu, Tian,\ \BBA\ Zhou}{Fan et~al.}{2019}]{Fan2019}
Fan, W., Lu, P., Tian, C., \BBA\ Zhou, J. \BBOP2019\BBCP.
\newblock \BBOQ Deducing certain fixes to graphs\BBCQ.
\newblock \lowercase{\BVOL}~12, \BPG\ 752–765. VLDB Endowment.

\bibitem[\protect\BCAY{Feuillade, Herzig,\ \BBA\ Rantsoudis}{Feuillade
  et~al.}{2020}]{FeuilladeHR20}
Feuillade, G., Herzig, A., \BBA\ Rantsoudis, C. \BBOP2020\BBCP.
\newblock \BBOQ Knowledge base repair: From active integrity constraints to
  active tboxes\BBCQ\
\newblock In Borgwardt, S.\BBACOMMA\  \BBA\ Meyer, T.\BEDS, {\Bem Proceedings
  of the 33rd International Workshop on Description Logics {(DL} 2020)
  co-located with the 17th International Conference on Principles of Knowledge
  Representation and Reasoning {(KR} 2020), Online Event [Rhodes, Greece],
  September 12th to 14th, 2020}, \lowercase{\BVOL}\ 2663 of {\Bem {CEUR}
  Workshop Proceedings}. CEUR-WS.org.

\bibitem[\protect\BCAY{Gad-Elrab, Stepanova, Urbani,\ \BBA\ Weikum}{Gad-Elrab
  et~al.}{2019}]{Gad-Elrab2019}
Gad-Elrab, M.~H., Stepanova, D., Urbani, J., \BBA\ Weikum, G. \BBOP2019\BBCP.
\newblock \BBOQ {Tracy: Tracing Facts over Knowledge Graphs and Text}\BBCQ\
\newblock In {\Bem WWW Conference}, \BPGS\ 3516--3520, New York, NY, USA. ACM.

\bibitem[\protect\BCAY{Geerts, Mecca, Papotti,\ \BBA\ Santoro}{Geerts
  et~al.}{2020}]{Geerts2020}
Geerts, F., Mecca, G., Papotti, P., \BBA\ Santoro, D. \BBOP2020\BBCP.
\newblock \BBOQ {Cleaning data with Llunatic}\BBCQ\
\newblock {\Bem The VLDB Journal}, {\Bem 29\/}(4), 867--892.

\bibitem[\protect\BCAY{Heist, Hertling, Ringler,\ \BBA\ Paulheim}{Heist
  et~al.}{2020}]{heist2020knowledge}
Heist, N., Hertling, S., Ringler, D., \BBA\ Paulheim, H. \BBOP2020\BBCP.
\newblock \BBOQ Knowledge graphs on the web-an overview.\BBCQ\
\newblock {\Bem {Knowledge Graphs for eXplainable Artificial Intelligence}},
  {\Bem 47}, 3--22.

\bibitem[\protect\BCAY{Heyvaert, {De Meester}, Dimou,\ \BBA\ Verborgh}{Heyvaert
  et~al.}{2019}]{Heyvaert2019}
Heyvaert, P., {De Meester}, B., Dimou, A., \BBA\ Verborgh, R. \BBOP2019\BBCP.
\newblock \BBOQ {Rule-driven inconsistency resolution for knowledge graph
  generation rules}\BBCQ\
\newblock {\Bem Semantic Web}, {\Bem 10\/}(6), 1071--1086.

\bibitem[\protect\BCAY{Hogan, Blomqvist, Cochez, D'amato, Melo, Gutierrez,
  Kirrane, Gayo, Navigli, Neumaier, Ngomo, Polleres, Rashid, Rula,
  Schmelzeisen, Sequeda, Staab,\ \BBA\ Zimmermann}{Hogan
  et~al.}{2022}]{Hogan2022}
Hogan, A., Blomqvist, E., Cochez, M., D'amato, C., Melo, G.~D., Gutierrez, C.,
  Kirrane, S., Gayo, J. E.~L., Navigli, R., Neumaier, S., Ngomo, A.-C.~N.,
  Polleres, A., Rashid, S.~M., Rula, A., Schmelzeisen, L., Sequeda, J., Staab,
  S., \BBA\ Zimmermann, A. \BBOP2022\BBCP.
\newblock \BBOQ {Knowledge Graphs}\BBCQ\
\newblock {\Bem ACM Computing Surveys}, {\Bem 54\/}(4), 1--37.

\bibitem[\protect\BCAY{Hohenecker\ \BBA\ Lukasiewicz}{Hohenecker\ \BBA\
  Lukasiewicz}{2020}]{Hohenecker2020}
Hohenecker, P.\BBACOMMA\  \BBA\ Lukasiewicz, T. \BBOP2020\BBCP.
\newblock \BBOQ {Ontology Reasoning with Deep Neural Networks}\BBCQ\
\newblock {\Bem Journal of Artificial Intelligence Research}, {\Bem 68},
  503--540.

\bibitem[\protect\BCAY{Horridge, Parsia,\ \BBA\ Sattler}{Horridge
  et~al.}{2009}]{HorridgeExplanations2009}
Horridge, M., Parsia, B., \BBA\ Sattler, U. \BBOP2009\BBCP.
\newblock \BBOQ Explaining inconsistencies in {OWL} ontologies\BBCQ\
\newblock In Godo, L.\BBACOMMA\  \BBA\ Pugliese, A.\BEDS, {\Bem Scalable
  Uncertainty Management}, \BPGS\ 124--137, Berlin, Heidelberg. Springer Berlin
  Heidelberg.

\bibitem[\protect\BCAY{Huang, van Harmelen,\ \BBA\ ten Teije}{Huang
  et~al.}{2005}]{huang2005reasoning}
Huang, Z., van Harmelen, F., \BBA\ ten Teije, A. \BBOP2005\BBCP.
\newblock \BBOQ Reasoning with inconsistent ontologies\BBCQ\
\newblock In Kaelbling, L.~P.\BBACOMMA\  \BBA\ Saffiotti, A.\BEDS, {\Bem
  IJCAI-05, Proceedings of the Nineteenth International Joint Conference on
  Artificial Intelligence, Edinburgh, Scotland, UK, July 30 - August 5, 2005},
  \BPGS\ 454--459. Professional Book Center.

\bibitem[\protect\BCAY{Jaeger}{Jaeger}{1994}]{jaeger_probabilistic_1994}
Jaeger, M. \BBOP1994\BBCP.
\newblock \BBOQ Probabilistic {Reasoning} in {Terminological} {Logics}\BBCQ\
\newblock In {\Bem KR'94}. Morgan Kaufmann.

\bibitem[\protect\BCAY{Ji, Pan, Cambria, Marttinen,\ \BBA\ Yu}{Ji
  et~al.}{2022}]{Ji2022}
Ji, S., Pan, S., Cambria, E., Marttinen, P., \BBA\ Yu, P.~S. \BBOP2022\BBCP.
\newblock \BBOQ {A Survey on Knowledge Graphs: Representation, Acquisition, and
  Applications}\BBCQ\
\newblock {\Bem IEEE Transactions on Neural Networks and Learning Systems},
  {\Bem 33\/}(2), 494--514.

\bibitem[\protect\BCAY{Jung\ \BBA\ Lutz}{Jung\ \BBA\
  Lutz}{2012}]{jung_ontology-based_2012}
Jung, J.~C.\BBACOMMA\  \BBA\ Lutz, C. \BBOP2012\BBCP.
\newblock \BBOQ Ontology-{Based} {Access} to {Probabilistic} {Data} with {OWL}
  {QL}\BBCQ.
\newblock Springer.

\bibitem[\protect\BCAY{Junker}{Junker}{2004}]{junker2004preferred}
Junker, U. \BBOP2004\BBCP.
\newblock \BBOQ Preferred explanations and relaxations for over-constrained
  problems\BBCQ\
\newblock In {\Bem AAAI-2004}.

\bibitem[\protect\BCAY{Kamide}{Kamide}{2013}]{kamide_comparison_2013}
Kamide, N. \BBOP2013\BBCP.
\newblock \BBOQ A {Comparison} of {Paraconsistent} {Description} {Logics}\BBCQ\
\newblock {\Bem International Journal of Intelligence Science}, {\Bem
  03\/}(02).

\bibitem[\protect\BCAY{Lambrix}{Lambrix}{2023}]{Lambrix2023}
Lambrix, P. \BBOP2023\BBCP.
\newblock \BBOQ {Completing and Debugging Ontologies: State-of-the-art and
  Challenges in Repairing Ontologies}\BBCQ\
\newblock {\Bem Journal of Data and Information Quality}, {\Bem 15\/}(4).

\bibitem[\protect\BCAY{Lao\ \BBA\ Cohen}{Lao\ \BBA\
  Cohen}{2010}]{10.1145/1835804.1835916}
Lao, N.\BBACOMMA\  \BBA\ Cohen, W.~W. \BBOP2010\BBCP.
\newblock \BBOQ Fast query execution for retrieval models based on
  path-constrained random walks\BBCQ\
\newblock In {\Bem Proceedings of the 16th ACM SIGKDD International Conference
  on Knowledge Discovery and Data Mining}, KDD '10, \BPG\ 881–888, New York,
  NY, USA. Association for Computing Machinery.

\bibitem[\protect\BCAY{Lembo, Lenzerini, Rosati, Ruzzi,\ \BBA\ Savo}{Lembo
  et~al.}{2010}]{lembo_inconsistency-tolerant_2010}
Lembo, D., Lenzerini, M., Rosati, R., Ruzzi, M., \BBA\ Savo, D.~F.
  \BBOP2010\BBCP.
\newblock \BBOQ Inconsistency-{Tolerant} {Semantics} for {Description}
  {Logics}\BBCQ\
\newblock In {\Bem Web {Reasoning} and {Rule} {Systems}}. Springer.

\bibitem[\protect\BCAY{Lembo, Lenzerini, Rosati, Ruzzi,\ \BBA\ Savo}{Lembo
  et~al.}{2015}]{lembo_inconsistency-tolerant_2015}
Lembo, D., Lenzerini, M., Rosati, R., Ruzzi, M., \BBA\ Savo, D.~F.
  \BBOP2015\BBCP.
\newblock \BBOQ Inconsistency-tolerant query answering in ontology-based data
  access\BBCQ\
\newblock {\Bem Journal of Web Semantics}, {\Bem 33}.

\bibitem[\protect\BCAY{Meilicke, Ruffinelli, Nolle, Paulheim,\ \BBA\
  Stuckenschmidt}{Meilicke et~al.}{2017}]{meilicke2017fast}
Meilicke, C., Ruffinelli, D., Nolle, A., Paulheim, H., \BBA\ Stuckenschmidt, H.
  \BBOP2017\BBCP.
\newblock \BBOQ Fast {ABox} consistency checking using incomplete reasoning and
  caching\BBCQ\
\newblock In {\Bem Rules and Reasoning: International Joint Conference, RuleML+
  RR 2017, London, UK, July 12--15, 2017, Proceedings 1}, \BPGS\ 168--183.
  Springer.

\bibitem[\protect\BCAY{Melo\ \BBA\ Paulheim}{Melo\ \BBA\
  Paulheim}{2020}]{Melo2020}
Melo, A.\BBACOMMA\  \BBA\ Paulheim, H. \BBOP2020\BBCP.
\newblock \BBOQ {Automatic detection of relation assertion errors and induction
  of relation constraints}\BBCQ\
\newblock {\Bem Semantic Web}, {\Bem 11\/}(5), 801--830.

\bibitem[\protect\BCAY{Mitchell, Cohen, Hruschka, Talukdar, Yang, Betteridge,
  Carlson, Dalvi, Gardner, Kisiel, et~al.}{Mitchell
  et~al.}{2018}]{mitchell2018never}
Mitchell, T., Cohen, W., Hruschka, E., Talukdar, P., Yang, B., Betteridge, J.,
  Carlson, A., Dalvi, B., Gardner, M., Kisiel, B., et~al. \BBOP2018\BBCP.
\newblock \BBOQ Never-ending learning\BBCQ\
\newblock {\Bem Communications of the ACM}, {\Bem 61\/}(5), 103--115.

\bibitem[\protect\BCAY{Musen}{Musen}{2015}]{Musen2015}
Musen, M.~A. \BBOP2015\BBCP.
\newblock \BBOQ {The prot{\'{e}}g{\'{e}} project}\BBCQ\
\newblock {\Bem AI Matters}, {\Bem 1\/}(4), 4--12.

\bibitem[\protect\BCAY{Nikitina, Rudolph,\ \BBA\ Glimm}{Nikitina
  et~al.}{2012}]{Nikitina2012}
Nikitina, N., Rudolph, S., \BBA\ Glimm, B. \BBOP2012\BBCP.
\newblock \BBOQ {Interactive ontology revision}\BBCQ\
\newblock {\Bem Journal of Web Semantics}, {\Bem 12-13\/}(721), 118--130.

\bibitem[\protect\BCAY{Paulheim}{Paulheim}{2016}]{Paulheim2016}
Paulheim, H. \BBOP2016\BBCP.
\newblock \BBOQ {Knowledge graph refinement: A survey of approaches and
  evaluation methods}\BBCQ\
\newblock {\Bem Semantic Web}, {\Bem 8\/}(3), 489--508.

\bibitem[\protect\BCAY{Paulheim\ \BBA\ Bizer}{Paulheim\ \BBA\
  Bizer}{2014}]{Paulheim2014}
Paulheim, H.\BBACOMMA\  \BBA\ Bizer, C. \BBOP2014\BBCP.
\newblock \BBOQ {Improving the Quality of Linked Data Using Statistical
  Distributions}\BBCQ\
\newblock {\Bem Int. J. Semant. Web Inf.}, {\Bem 10\/}(2), 63--86.

\bibitem[\protect\BCAY{Paulheim\ \BBA\ Stuckenschmidt}{Paulheim\ \BBA\
  Stuckenschmidt}{2016}]{paulheim2016fast}
Paulheim, H.\BBACOMMA\  \BBA\ Stuckenschmidt, H. \BBOP2016\BBCP.
\newblock \BBOQ Fast approximate a-box consistency checking using machine
  learning\BBCQ\
\newblock In {\Bem The Semantic Web. Latest Advances and New Domains: 13th
  International Conference, ESWC 2016, Heraklion, Crete, Greece, May 29--June
  2, 2016, Proceedings 13}, \BPGS\ 135--150. Springer.

\bibitem[\protect\BCAY{{Pellissier Tanon}, Bourgaux,\ \BBA\
  Suchanek}{{Pellissier Tanon} et~al.}{2019}]{PellissierTanon2019}
{Pellissier Tanon}, T., Bourgaux, C., \BBA\ Suchanek, F. \BBOP2019\BBCP.
\newblock \BBOQ {Learning How to Correct a Knowledge Base from the Edit
  History}\BBCQ\
\newblock In {\Bem ACM Web Conference}, \BPGS\ 1465--1475, New York, NY, USA.
  ACM.

\bibitem[\protect\BCAY{{Pellissier Tanon}\ \BBA\ Suchanek}{{Pellissier Tanon}\
  \BBA\ Suchanek}{2021}]{PellissierTanon2021}
{Pellissier Tanon}, T.\BBACOMMA\  \BBA\ Suchanek, F. \BBOP2021\BBCP.
\newblock \BBOQ {Neural Knowledge Base Repairs}\BBCQ\
\newblock In {\Bem ESWC}, \lowercase{\BVOL}\ 12731 LNCS, \BPGS\ 287--303.

\bibitem[\protect\BCAY{{Pellissier Tanon}, Weikum,\ \BBA\ Suchanek}{{Pellissier
  Tanon} et~al.}{2020}]{PellissierTanon2020}
{Pellissier Tanon}, T., Weikum, G., \BBA\ Suchanek, F. \BBOP2020\BBCP.
\newblock \BBOQ {YAGO 4: A Reason-able Knowledge Base}\BBCQ\
\newblock In {\Bem Lecture Notes in Computer Science (including subseries
  Lecture Notes in Artificial Intelligence and Lecture Notes in
  Bioinformatics)}, \lowercase{\BVOL}\ 12123 LNCS, \BPGS\ 583--596.

\bibitem[\protect\BCAY{Pe{\~{n}}aloza}{Pe{\~{n}}aloza}{2019}]{Penaloza2019}
Pe{\~{n}}aloza, R. \BBOP2019\BBCP.
\newblock \BBOQ {Making Decisions with Knowledge Base Repairs}\BBCQ\
\newblock In {\Bem Lecture Notes in Computer Science (including subseries
  Lecture Notes in Artificial Intelligence and Lecture Notes in
  Bioinformatics)}, \lowercase{\BVOL}\ 11676 LNAI, \BPGS\ 259--271.

\bibitem[\protect\BCAY{Pensel\ \BBA\ Turhan}{Pensel\ \BBA\
  Turhan}{2018}]{Pensel2018}
Pensel, M.\BBACOMMA\  \BBA\ Turhan, A.-Y. \BBOP2018\BBCP.
\newblock \BBOQ Reasoning in the defeasible description logic — computing
  standard inferences under rational and relevant semantics\BBCQ\
\newblock {\Bem International Journal of Approximate Reasoning}, {\Bem 103},
  28–70.

\bibitem[\protect\BCAY{Qi, Pan,\ \BBA\ Ji}{Qi
  et~al.}{2007}]{qi_possibilistic_2007}
Qi, G., Pan, J.~Z., \BBA\ Ji, Q. \BBOP2007\BBCP.
\newblock \BBOQ A possibilistic extension of description logics\BBCQ\
\newblock In {\Bem {CEUR} {Workshop} {Proceedings}}, \lowercase{\BVOL}\ 250.
  CEUR-WS.
\newblock ISSN: 1613-0073.

\bibitem[\protect\BCAY{Qi\ \BBA\ Yang}{Qi\ \BBA\ Yang}{2008}]{Qi2008}
Qi, G.\BBACOMMA\  \BBA\ Yang, F. \BBOP2008\BBCP.
\newblock \BBOQ {A Survey of Revision Approaches in Description Logics}\BBCQ\
\newblock In {\Bem Lecture Notes in Computer Science (including subseries
  Lecture Notes in Artificial Intelligence and Lecture Notes in
  Bioinformatics)}, \lowercase{\BVOL}\ 5341 LNCS, \BPGS\ 74--88.

\bibitem[\protect\BCAY{Reiter}{Reiter}{1987}]{reiter1987theory}
Reiter, R. \BBOP1987\BBCP.
\newblock \BBOQ A theory of diagnosis from first principles\BBCQ\
\newblock {\Bem Artificial intelligence}, {\Bem 32\/}(1), 57--95.

\bibitem[\protect\BCAY{Rodler}{Rodler}{2016}]{Rodler2016}
Rodler, P. \BBOP2016\BBCP.
\newblock \BBOQ {Interactive Debugging of Knowledge Bases}\BBCQ.

\bibitem[\protect\BCAY{Rodler}{Rodler}{2022}]{Rodler2022}
Rodler, P. \BBOP2022\BBCP.
\newblock \BBOQ {One step at a time: An efficient approach to query-based
  ontology debugging}\BBCQ\
\newblock {\Bem Knowledge-Based Systems}, {\Bem 251}, 108987.

\bibitem[\protect\BCAY{Rosati}{Rosati}{2011}]{rosati_complexity_2011}
Rosati, R. \BBOP2011\BBCP.
\newblock \BBOQ On the complexity of dealing with inconsistency in description
  logic ontologies\BBCQ\
\newblock In {\Bem Proceedings of the {Twenty}-{Second} international joint
  conference on {Artificial} {Intelligence} - {Volume} {Volume} {Two}},
  {IJCAI}'11, Barcelona, Catalonia, Spain. AAAI Press.

\bibitem[\protect\BCAY{Rosati, Ruzzi, Graziosi,\ \BBA\ Masotti}{Rosati
  et~al.}{2012}]{rosati_evaluation_2012}
Rosati, R., Ruzzi, M., Graziosi, M., \BBA\ Masotti, G. \BBOP2012\BBCP.
\newblock \BBOQ Evaluation of {Techniques} for {Inconsistency} {Handling} in
  {OWL} 2 {QL} {Ontologies}\BBCQ\
\newblock In {\Bem {ISWC}'12}. Springer.

\bibitem[\protect\BCAY{Schekotihin, Rodler,\ \BBA\ Schmid}{Schekotihin
  et~al.}{2018}]{Schekotihin2018}
Schekotihin, K., Rodler, P., \BBA\ Schmid, W. \BBOP2018\BBCP.
\newblock \BBOQ {Ontodebug: Interactive ontology debugging plug-in for
  prot{\'{e}}g{\'{e}}}\BBCQ\
\newblock {\Bem Lecture Notes in Computer Science (including subseries Lecture
  Notes in Artificial Intelligence and Lecture Notes in Bioinformatics)}, {\Bem
  10833 LNCS\/}(January), 340--359.

\bibitem[\protect\BCAY{Senaratne}{Senaratne}{2023}]{10.1145/3543873.3587536}
Senaratne, A. \BBOP2023\BBCP.
\newblock \BBOQ {SEKA}: Seeking knowledge graph anomalies\BBCQ\
\newblock In {\Bem Companion Proceedings of the ACM Web Conference 2023}, WWW
  '23 Companion, \BPG\ 568–572, New York, NY, USA. Association for Computing
  Machinery.

\bibitem[\protect\BCAY{Shchekotykhin, Friedrich, Fleiss,\ \BBA\
  Rodler}{Shchekotykhin et~al.}{2012}]{Shchekotykhin2012}
Shchekotykhin, K., Friedrich, G., Fleiss, P., \BBA\ Rodler, P. \BBOP2012\BBCP.
\newblock \BBOQ {Interactive ontology debugging: Two query strategies for
  efficient fault localization}\BBCQ\
\newblock {\Bem Journal of Web Semantics}, {\Bem 12-13}, 88--103.

\bibitem[\protect\BCAY{Shchekotykhin, Friedrich, Rodler,\ \BBA\
  Fleiss}{Shchekotykhin et~al.}{2014}]{shchekotykhin2014sequential}
Shchekotykhin, K., Friedrich, G., Rodler, P., \BBA\ Fleiss, P. \BBOP2014\BBCP.
\newblock \BBOQ Sequential diagnosis of high cardinality faults in
  knowledge-bases by direct diagnosis generation\BBCQ\
\newblock In {\Bem ECAI 2014}, \BPGS\ 813--818. IOS Press.

\bibitem[\protect\BCAY{Suntisrivaraporn, Qi, Ji,\ \BBA\ Haase}{Suntisrivaraporn
  et~al.}{2008}]{suntisrivaraporn2008modularization}
Suntisrivaraporn, B., Qi, G., Ji, Q., \BBA\ Haase, P. \BBOP2008\BBCP.
\newblock \BBOQ A modularization-based approach to finding all justifications
  for {OWL DL} entailments\BBCQ\
\newblock In {\Bem The Semantic Web: 3rd Asian Semantic Web Conference, ASWC
  2008, Bangkok, Thailand, December 8-11, 2008. Proceedings. 3}, \BPGS\ 1--15.
  Springer.

\bibitem[\protect\BCAY{T{\"{o}}pper, Knuth,\ \BBA\ Sack}{T{\"{o}}pper
  et~al.}{2012}]{Topper2012}
T{\"{o}}pper, G., Knuth, M., \BBA\ Sack, H. \BBOP2012\BBCP.
\newblock \BBOQ {DBpedia ontology enrichment for inconsistency detection}\BBCQ\
\newblock In {\Bem I-SEMANTICS '12}, \BPG~33. ACM Press.

\bibitem[\protect\BCAY{Tran, Gad-Elrab, Stepanova, Kharlamov,\ \BBA\
  Str{\"o}tgen}{Tran et~al.}{2020}]{tran2020fast}
Tran, T.-K., Gad-Elrab, M.~H., Stepanova, D., Kharlamov, E., \BBA\
  Str{\"o}tgen, J. \BBOP2020\BBCP.
\newblock \BBOQ Fast computation of explanations for inconsistency in
  large-scale knowledge graphs\BBCQ\
\newblock In {\Bem Proceedings of The Web Conference 2020}, \BPGS\ 2613--2619.

\bibitem[\protect\BCAY{Trivela, Stoilos,\ \BBA\ Vassalos}{Trivela
  et~al.}{2018}]{trivela_framework_2018}
Trivela, D., Stoilos, G., \BBA\ Vassalos, V. \BBOP2018\BBCP.
\newblock \BBOQ A {Framework} and {Positive} {Results} for
  {IAR}-answering\BBCQ\
\newblock {\Bem Proceedings of the AAAI Conference on Artificial Intelligence},
  {\Bem 32\/}(1).
\newblock Number: 1.

\bibitem[\protect\BCAY{Trivela, Stoilos,\ \BBA\ Vassalos}{Trivela
  et~al.}{2019}]{trivela_query_2019}
Trivela, D., Stoilos, G., \BBA\ Vassalos, V. \BBOP2019\BBCP.
\newblock \BBOQ Query {Rewriting} for {DL} {Ontologies} {Under} the {ICAR}
  {Semantics}\BBCQ\
\newblock In {\Bem {RuleML}+{RR} 2019}. Springer-Verlag.

\bibitem[\protect\BCAY{Tsalapati, Stoilos, Stamou,\ \BBA\ Koletsos}{Tsalapati
  et~al.}{2016}]{tsalapati_efficient_2016}
Tsalapati, E., Stoilos, G., Stamou, G., \BBA\ Koletsos, G. \BBOP2016\BBCP.
\newblock \BBOQ Efficient query answering over expressive inconsistent
  description logics\BBCQ.
\newblock {IJCAI}'16.

\bibitem[\protect\BCAY{Wang, Yan, Wang, Jia, Zhang, Zhang,\ \BBA\ Wang}{Wang
  et~al.}{2018}]{wang2018acekg}
Wang, R., Yan, Y., Wang, J., Jia, Y., Zhang, Y., Zhang, W., \BBA\ Wang, X.
  \BBOP2018\BBCP.
\newblock \BBOQ Acekg: A large-scale knowledge graph for academic data
  mining\BBCQ\
\newblock In {\Bem Proceedings of the 27th ACM international conference on
  information and knowledge management}, \BPGS\ 1487--1490.

\bibitem[\protect\BCAY{Xue\ \BBA\ Zou}{Xue\ \BBA\ Zou}{2022}]{Xue2022}
Xue, B.\BBACOMMA\  \BBA\ Zou, L. \BBOP2022\BBCP.
\newblock \BBOQ {Knowledge Graph Quality Management: a Comprehensive
  Survey}\BBCQ\
\newblock In {\Bem IEEE Transactions on Knowledge and Data Engineering},
  \lowercase{\BVOL}~14, \BPGS\ 1--1. IEEE.

\bibitem[\protect\BCAY{Ye, Xu, Zhang, Wu,\ \BBA\ Dai}{Ye et~al.}{2023}]{Ye2023}
Ye, C., Xu, H., Zhang, H., Wu, Y., \BBA\ Dai, G. \BBOP2023\BBCP.
\newblock \BBOQ {Grier: graph repairing based on iterative embedding and
  rules}\BBCQ\
\newblock {\Bem Knowledge and Information Systems}, {\Bem 65\/}(8), 3273--3294.

\bibitem[\protect\BCAY{Zhang, Xiao, Lin,\ \BBA\ Van~den Bussche}{Zhang
  et~al.}{2014}]{zhang_inconsistency-tolerant_2014}
Zhang, X., Xiao, G., Lin, Z., \BBA\ Van~den Bussche, J. \BBOP2014\BBCP.
\newblock \BBOQ Inconsistency-tolerant reasoning with {OWL} {DL}\BBCQ\
\newblock {\Bem International Journal of Approximate Reasoning}, {\Bem
  55\/}(2).

\end{thebibliography}

\end{document}